\pdfoutput=1
\documentclass[10pt]{article}

\usepackage[utf8]{inputenc}
\usepackage[T1]{fontenc}

\usepackage[scale=1.0]{XCharter}
\usepackage[letterpaper,top=1in,bottom=1in]{geometry}
\usepackage[final]{microtype}
\usepackage[english]{babel}
\usepackage[parfill]{parskip}
\usepackage{graphicx}
\usepackage{amsmath,amsfonts,amssymb,amsthm}
\usepackage{booktabs,multirow,multicol}
\usepackage{array}
\usepackage{wrapfig,needspace}
\usepackage[labelfont=bf,font=footnotesize,width=\textwidth]{caption}
\usepackage[format=hang]{subcaption}
\usepackage[table]{xcolor}
\usepackage{enumitem}
\usepackage{url}
\usepackage{fontawesome5}
\usepackage{authblk}
\usepackage{tcolorbox}
\tcbuselibrary{skins}

\definecolor{mydarkblue}{rgb}{0,0.08,0.45}
\usepackage[colorlinks=true,linktoc=all,hyperfootnotes=false]{hyperref}
\hypersetup{
  linkcolor=mydarkblue, citecolor=mydarkblue, urlcolor=mydarkblue,
  pdftitle={Latent Prior GANs: Efficient Decoder Refinement via Inherited Generative Structure},
  pdfauthor={Wenjie Shu, Xianghong Fang, Dehan Kong, Tim G. J. Rudner}
}
\usepackage[nameinlink,capitalize,noabbrev]{cleveref}
\crefformat{equation}{(#2#1#3)}
\usepackage{csquotes}
\usepackage[%
minnames=1,maxnames=99,maxcitenames=2,
style=alphabetic,
doi=false,
url=false,
giveninits=true,
hyperref,
natbib,
backend=bibtex,
sorting=nyt,
backref=true
]{biblatex}%
\renewbibmacro{in:}{%
  \ifentrytype{article}{}{\printtext{\bibstring{in}\intitlepunct}}}

\renewbibmacro*{journal}{%
  \iffieldundef{journaltitle}
    {}
    {\printtext[journaltitle]{%
       \printfield[noformat]{journaltitle}%
       \setunit{\subtitlepunct}%
       \printfield[noformat]{journalsubtitle}}}}

\DeclareFieldFormat{sentencecase}{\MakeSentenceCase*{#1}}

\renewbibmacro*{title}{%
  \ifthenelse{\iffieldundef{title}\AND\iffieldundef{subtitle}}
    {}
    {\ifthenelse{\ifentrytype{article}\OR\ifentrytype{inbook}%
      \OR\ifentrytype{incollection}\OR\ifentrytype{inproceedings}%
      \OR\ifentrytype{inreference}\OR\ifentrytype{misc}}
      {\printtext[title]{%
        \printfield[sentencecase]{title}%
        \setunit{\subtitlepunct}%
        \printfield[sentencecase]{subtitle}}}%
      {\printtext[title]{%
        \printfield[titlecase]{title}%
        \setunit{\subtitlepunct}%
        \printfield[titlecase]{subtitle}}}%
     \newunit}%
  \printfield{titleaddon}}

\AtEveryBibitem{%
\ifentrytype{article}{
    \clearfield{urldate}%
    \clearfield{eprint}
    \clearfield{eid}
}{}
\ifentrytype{book}{
    \clearfield{url}%
    \clearfield{urldate}%
    \clearfield{eprint}
}{}
\ifentrytype{collection}{
    \clearfield{url}%
    \clearfield{urldate}%
    \clearfield{eprint}
}{}
\ifentrytype{incollection}{
    \clearfield{url}%
    \clearfield{urldate}%
    \clearfield{eprint}
}{}
}

\AtEveryBibitem{
    \clearfield{pages}
    \clearfield{review}%
    \clearfield{series}
    \clearfield{volume}
    \clearfield{month}
    \clearfield{isbn}
    \clearfield{issn}
    \clearlist{location}
    \clearfield{series}
    \clearlist{publisher}
    \clearname{editor}
}{}

\newcommand{\wraptopalign}{%
  \ifdim\pagetotal=0pt\vspace{0pt}\else\vspace{-\intextsep}\fi
}

\newcolumntype{x}[1]{>{\centering\arraybackslash}p{#1pt}}
\newcolumntype{y}[1]{>{\raggedright\arraybackslash}p{#1pt}}

\title{Latent Prior GANs: Adversarial Learning for Targeted\\Generation-Time Decoder Finetuning}
\author{
  {\normalfont\bfseries
  Wenjie Shu$^{1,*}$ \quad
  Xianghong Fang$^{1,*,{\dagger}}$ \quad
  Dehan Kong$^{1}$ \quad
  Tim G. J. Rudner$^{1,2}$
  } \\
  {\normalfont
  \vspace*{10pt}
  $^1$University of Toronto \quad $^2$Vijil
  } \\
  {\normalfont
  \vspace*{15pt}
  \begin{center}
    \href{https://wenjieshu.github.io/Latent-Prior-GAN/}{\faGlobe\enspace Website} \quad
    \href{https://github.com/WenjieShu/Latent-Prior-GAN}{\faGithub\enspace Code \& Models}
  \end{center}
  }
}

\date{}

\begin{document}
\maketitle

\begingroup
\renewcommand\thefootnote{}
\footnotetext{$^{*}$Equal contribution.}
\footnotetext{$^{\dagger}$Corresponding author: \href{mailto:xianghong@cs.toronto.edu}{xianghong.fang@mail.utoronto.ca}}
\endgroup

\begin{abstract}
Modern latent generative models learn strong generative structure through stable, non-adversarial training objectives in their latent generators. Their success changes a key premise of adversarial generation: this structure need no longer be learned from scratch through adversarial optimization, but can instead be inherited from a pretrained model. We refer to this inherited generative structure, comprising the learned generation-time latent distribution and pretrained latent-to-image decoding knowledge, as the latent prior. We introduce Latent Prior GANs (LP-GANs), which repurpose adversarial learning from discovering generative structure to targeted generation-time decoder finetuning. LP-GANs freezes the pretrained latent generator and adversarially optimizes its decoder on generation-time latents, directly correcting the decoder train-generation mismatch: conventional decoders are optimized on reconstruction latents but deployed on generated latents. The reconstruction and perceptual losses require paired images, which generated latents do not provide, so we construct intermediate latents that still correspond to a real image. We show that across four latent generators of varying size, five epochs of adversarial adaptation are enough to improve generation quality over the pretrained model. We hope this makes adversarial training cheaper when strong pretrained models are already available.
\end{abstract}

\section{Introduction}
\label{sec:introduction}
\begin{figure}[t!]
    \centering
    \includegraphics[width=\linewidth]{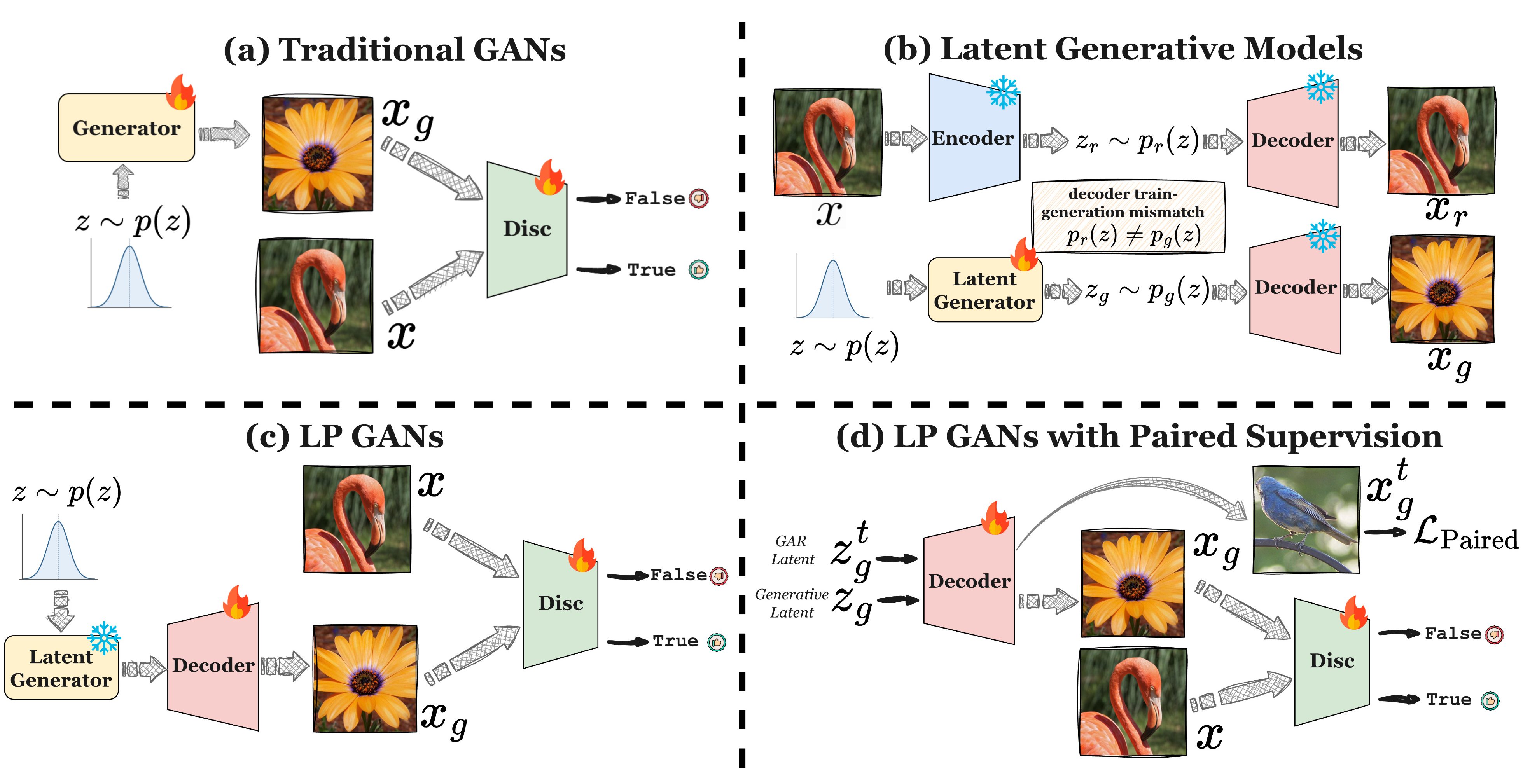}
    \vspace{-14pt}
    \caption{\small{\textbf{Overview and motivation of Latent Prior GANs (LP-GANs).}
    \textbf{(a)} Conventional GANs couple generative structure discovery and refinement within a training-intensive adversarial process.
    \textbf{(b)} Pretrained latent generative models provide learned generative structure, but their decoders are optimized on reconstruction latents and deployed on generated latents, creating a decoder train-generation mismatch.
    \textbf{(c)} LP-GANs inherit this latent prior, freeze the latent generator, and correct this mismatch through targeted, generation-time decoder refinement.
    \textbf{(d)} Generation-aware intermediate latents provide paired supervision that anchors decoder adaptation toward the generation-time latent distribution.
    }}
    \label{fig:introduction}
\vspace{-8pt}
\end{figure}

Generative adversarial networks (GANs) directly optimize generated samples against the data distribution and have been highly effective for high-fidelity image synthesis~\citep{Goodfellow2014GenerativeAN,Brock2018LargeSG,Karras2019ASG,Karras2020AnalyzingAI,Karras2021AliasFreeGA}. However, GANs are also notoriously difficult and costly to train. In large part, this is because a from-scratch GAN is asked to do two things at once: find the structure that makes synthesis possible, and turn that structure into high-quality samples. In practice, both tasks fall to the same adversarial process, which is why large-scale GAN training is difficult to optimize (\Cref{fig:introduction}(a)).

Rather than learning such structure from scratch, one can start from generative structure learned beforehand, as already offered by pretrained latent generative models. We therefore call their inherited structure the \emph{latent prior}. It contains what the latent generator has learned about its generation-time latent distribution, together with the latent-to-image decoding knowledge in the pretrained decoder~\citep{Esser2021TamingTF,Rombach2022HighResolutionIS,Peebles2022ScalableDM,Geng2025MeanFF,Geng2025ImprovedMF}. Adversarial training can then start from this prior, instead of paying the full cost of discovering such generative structure from scratch.

However, this latent prior contains an internal mismatch: its two components were never trained to match each other. The two are learned in separate stages~\citep{Esser2021TamingTF,Rombach2022HighResolutionIS}. During tokenizer training, the decoder sees reconstruction latents $z_r \sim p_r(z)$ from the encoder. At generation time the latents instead come from the generator, $z_g \sim p_g(z)$. The generator models $p_r$ only approximately, so $p_g \neq p_r$ in general and the decoder is applied off the distribution under which it was optimized. We refer to this discrepancy as the \emph{decoder train-generation mismatch}~\citep{Fang2026GenerationAwareRecon} (\Cref{fig:introduction}(b)). And $p_g$ is already the distribution used at deployment, so the mismatch left is at the decoder, which has never been optimized on it.

This raises a natural question:
\vspace*{2pt}
\begin{tcolorbox}[
  enhanced,
  colback=blue!4,
  colframe=blue!55!black,
  leftrule=3pt,
  toprule=0.4pt,
  bottomrule=0.4pt,
  rightrule=0.4pt,
  arc=2pt,
  boxsep=1.5pt,
  left=8pt, right=8pt, top=4pt, bottom=4pt,
  width=\linewidth,
  before skip=2pt,
  after skip=5pt,
]
\begin{center}
{Can adversarial optimization be repurposed from learning generative structure\\to finetuning a pretrained decoder for latents encountered at generation time?}
\end{center}
\end{tcolorbox}
\vspace*{2pt}

We answer this question with \textbf{Latent Prior GANs (LP-GANs)}, shown in \Cref{fig:introduction}(c). In LP-GANs, the latent generator already defines $p_g(z)$, the distribution encountered at generation time, so we keep it fixed and perform adversarial optimization on only the decoder. The decoder is trained on latents drawn from $p_g(z)$, which changes how that distribution is decoded rather than changing the distribution itself.

Fully generated latents differ from reconstruction latents in one important way: they have no paired ground-truth images, so the correspondence needed for perceptual supervision is missing. To address this, we use intermediate latents along a trajectory from reconstruction to generation~\citep{Fang2026GenerationAwareRecon} that move progressively toward generation-time latents while remaining paired with their source images. This \emph{generation-aware paired supervision} anchors the decoder through paired reconstruction and perceptual objectives~\citep{Johnson2016PerceptualLF,Zhang2018TheUE} while adversarial learning optimizes the decoder on fully generated latents (\Cref{fig:introduction}(d)).

Strikingly, we find that, across four latent generators of varying sizes, five epochs of adversarial adaptation are enough to improve generation quality, at a small fraction of the cost of learning structure from scratch.

Our main contributions are:
\begin{enumerate}[topsep=0pt, align=left, leftmargin=15pt, labelindent=1pt,
listparindent=\parindent, labelwidth=0pt, itemindent=!, itemsep=3pt, parsep=0pt]

\item We introduce \textbf{Latent Prior GANs (LP-GANs)}, which repurpose adversarial learning from discovering generative structure to targeted generation-time decoder finetuning.

\item We directly optimize the decoder under the fully generated latent distribution $p_g(z)$ while keeping the latent generator frozen, and use generation-aware intermediate latents to provide paired reconstruction and perceptual supervision that anchors this refinement.

\item Across four latent generators of varying sizes, LP-GANs improve generation quality within five epochs of adversarial adaptation, which makes decoder refinement over inherited structure both effective and cheap.
\end{enumerate}

\section{Related Work}
\label{sec:related_work}
\paragraph{Adversarial image generation.}
Generative adversarial networks, or GANs~\citep{Goodfellow2014GenerativeAN}, learn the image distribution by pitting two networks against each other. Later work made them easier to train and improved their samples, by changing the objective~\citep{Arjovsky2017WassersteinGA,Gulrajani2017ImprovedTO}, regularizing the discriminator~\citep{Miyato2018SpectralNF,Mescheder2018WhichTM}, or scaling up the architecture~\citep{Brock2018LargeSG,Karras2019ASG,Karras2020AnalyzingAI,Karras2021AliasFreeGA}. All of these GANs start from random noise. A single adversarial process discovers the generative structure required for synthesis and refines it into high-quality samples. LP-GANs removes the first of these responsibilities: generative structure is inherited as a latent prior from a pretrained latent generative model, leaving adversarial optimization to focus on refinement.

\paragraph{Adversarial post-training of pretrained generators.} Recent work applies adversarial post-training, or adversarial optimization of a pretrained generative model. One line uses adversarial or distribution-matching objectives primarily to compress sampling into few or single steps: ADD~\citep{Sauer2023AdversarialDD}, LADD~\citep{Sauer2024FastHI}, SiDA~\citep{Zhou2025AdversarialSI}, and SDXL-Lightning~\citep{Lin2024SDXLLightningPA} incorporate adversarial losses into diffusion distillation, while UFOGen~\citep{Xu2024UFOGenYF} initializes from a pretrained diffusion model to train a one-step diffusion-GAN hybrid and D2O~\citep{Zheng2025RevisitingDM} treats diffusion training as generative pre-training followed by lightweight GAN fine-tuning for one-step generation. Beyond distillation, ADT~\citep{Shen2025ADTTD} adversarially fine-tunes a pretrained generator to better align inference-time outputs with real data. 
These methods differ from LP-GANs in where subsequent optimization is applied. They modify, distill, or fine-tune the generative model or its sampling process, whereas LP-GANs keeps the latent generator entirely frozen and uses it only as a source of generation-time latents $z_g \sim p_g(z)$. The adversarial signal acts exclusively on the downstream decoder. LP-GANs thus aims not at faster sampling, but at aligning the decoder with a latent distribution the generator already produces.

\paragraph{Latent decoding and generation-time alignment.}
Latent generative models build the generative distribution over a learned representation, or latent space. The representation is discrete in VQ-VAE, VQ-VAE-2, and VQGAN~\citep{Oord2017NeuralDR,Razavi2019GeneratingDH,Esser2021TamingTF}, and continuous in latent diffusion and flow-based models~\citep{Rombach2022HighResolutionIS,Peebles2022ScalableDM,Geng2025MeanFF,Geng2025ImprovedMF}. Decoders in these pipelines are commonly trained with reconstruction, perceptual, and adversarial objectives~\citep{Esser2021TamingTF,Rombach2022HighResolutionIS}, but are typically optimized on reconstruction latents $z_r \sim p_r(z)$ and deployed on generated latents, giving rise to the decoder train-generation mismatch~\citep{Fang2026GenerationAwareRecon}.

Recent work narrows this gap by exposing the decoder to latents that move toward generation while preserving correspondence with source images. RobusTok~\citep{Qiu2025ImageTN} performs generator-aware decoder post-training, introducing for autoregressive generators a preservation ratio that retains a fraction of reconstruction tokens during sampling. Generation-aware reconstruction~\citep{Fang2026GenerationAwareRecon} constructs a reconstruction-to-generation trajectory and treats the resulting intermediate latents as a continuum from $p_r(z)$ toward $p_g(z)$ under paired supervision. In both cases, supervision stops short of fully generated latents from $p_g(z)$.

LP-GANs extends this generation-aware perspective to the generation-time distribution itself: it optimizes the decoder directly on fully generated latents, which admit no paired reconstruction target, and uses intermediate reconstruction-to-generation latents only to anchor this adaptation. The key distinction is therefore not adversarial decoder training, already standard in these pipelines, but the latent distribution under which it is applied.

\section{Preliminaries: Latent Generative Models}
\label{sec:preliminaries}
\paragraph{Two-stage latent generative modeling.}
We focus on diffusion- and flow-based latent generative models with continuous latent spaces~\citep{Rombach2022HighResolutionIS,Peebles2022ScalableDM,Geng2025MeanFF,Geng2025ImprovedMF}. These models typically adopt a two-stage formulation. The first stage learns a tokenizer, consisting of an encoder $\mathcal{E}_{\theta}$ and a decoder $\mathcal{D}_{\phi}$, which map an image $x \sim p_{\text{data}}$ to a latent representation $z_r = \mathcal{E}_{\theta}(x)$ and back to an image $\hat{x} = \mathcal{D}_{\phi}(z_r)$. The tokenizer is optimized with reconstruction and perceptual objectives~\citep{Johnson2016PerceptualLF,Zhang2018TheUE}, commonly together with an adversarial objective~\citep{Goodfellow2014GenerativeAN}:
\begin{equation}
\min_{\theta,\,\phi}\;
\mathbb{E}_{x \sim p_{\text{data}}}
\Big[
\underbrace{\mathcal{L}_{\text{rec}}\big(x,\hat{x}\big)
+ \mathcal{L}_{\text{perc}}\big(x,\hat{x}\big)}_{\text{paired objectives}}
\;+
\underbrace{\;\lambda\;\mathcal{L}_{\text{adv}}\big(\hat{x}\big)}_{\text{distributional objective}}
\Big],
\label{eq:stage1}
\end{equation}
where $\lambda$ balances the adversarial term against the paired terms, and $\hat{x} = \mathcal{D}_{\phi}(\mathcal{E}_{\theta}(x))$: in every term of Equation~\ref{eq:stage1}, the decoder receives only encoder-produced latents. The first two terms are \emph{paired}: they compare the reconstruction $\hat{x}$ with its corresponding source image $x$, either in pixel or feature space. The adversarial term is \emph{distributional} and requires no such image-level correspondence. The second stage then learns a latent generator $\mathcal{G}_{\varphi}$ over the latents produced by the frozen encoder, so that samples drawn from $\mathcal{G}_{\varphi}$ can be decoded by $\mathcal{D}_{\phi}$ into images. The discussion below does not depend on how $\mathcal{G}_{\varphi}$ is parameterized; it only needs access to its generation-time sampling procedure.

\paragraph{Reconstruction and generation latents.}
The two-stage latent generative pipeline exposes the decoder to distinct reconstruction and generation latent distributions. During first-stage training, the decoder receives \emph{reconstruction latents} produced by the encoder,
\begin{equation*}
z_r \sim p_r(z), \qquad
p_r = (\mathcal{E}_{\theta})_{\#}\, p_{\text{data}},
\end{equation*}
where $p_r$ is the pushforward of the data distribution through the encoder. In the second stage, the tokenizer is frozen and a latent generator $\mathcal{G}_{\varphi}$ is trained to approximately model $p_r$. At generation time, the decoder instead receives \emph{generation latents} produced by the trained latent generator through its deployed sampling procedure,
\begin{equation*}
z_g
=
\mathcal{S}(\mathcal{G}_{\varphi}, \epsilon, \xi)
\sim p_g(z),
\qquad
\epsilon \sim \mathcal{N}(0,I),
\quad
\xi \sim p_{\xi},
\end{equation*}
where $\epsilon$ denotes the noise initialization, $\xi$ denotes any additional randomness used by the sampler, and $p_g$ is the latent distribution induced by the deployed generation procedure.
Although $\mathcal{G}_{\varphi}$ is trained to model $p_r$, the resulting generation-time distribution need not match it exactly, so $p_g \neq p_r$ in general~\citep{Fang2026GenerationAwareRecon}.

\paragraph{Decoder train-generation mismatch.}
In Equation~\ref{eq:stage1}, all updates to the decoder are induced by reconstruction latents $z_r \sim p_r$, whereas at generation time the same decoder is applied to generation latents $z_g \sim p_g$. When $p_g \neq p_r$, this creates a \emph{decoder train-generation mismatch}~\citep{Fang2026GenerationAwareRecon}: the decoder is deployed on a latent distribution different from the one on which it was optimized. This mismatch cannot be addressed by simply replacing $z_r$ with $z_g$ in the first-stage objective, because a generated latent $z_g$ has no corresponding source image $x$, leaving the paired reconstruction and perceptual objectives undefined. The adversarial objective, by contrast, remains directly applicable under $p_g$ because it requires no image-level correspondence. Together, the decoder train-generation mismatch and this supervision asymmetry motivate LP-GANs, which apply adversarial supervision directly to generation-time latents $z_g \sim p_g$. We introduce LP-GANs next.

\section{Latent Prior GANs}
\label{sec:method}

We introduce \textbf{Latent Prior GANs (LP-GANs)}, which inherit the \emph{latent prior} provided by pretrained latent generative models and correct the decoder train-generation mismatch identified in Section~\ref{sec:preliminaries}.  LP-GANs directly applies adversarial supervision to the decoder under the fully generated latent distribution $p_g(z)$ while keeping the latent generator frozen (\Cref{sec:decoder_refinement}). 
Since fully generated latents admit no paired target, we use generation-aware intermediate latents to provide paired supervision that anchors this generation-time refinement (\Cref{sec:paired_supervision}).

\subsection{Generation-Time Decoder Refinement}
\label{sec:decoder_refinement}
LP-GANs optimizes the decoder on the latent distribution it encounters at deployment. Starting from a pretrained latent generator $\mathcal{G}_{\varphi}$ and decoder $\mathcal{D}_{\phi_0}$, we keep $\mathcal{G}_{\varphi}$ frozen and initialize $\mathcal{D}_{\phi}$ from $\mathcal{D}_{\phi_0}$. We sample
$z_g = \mathcal{S}(\mathcal{G}_{\varphi}, \xi) \sim p_g(z)$ with
$\xi \sim p_{\xi}$, decode it as $\mathcal{D}_{\phi}(z_g)$, and let an image-space discriminator $\mathcal{C}_{\omega}$ distinguish the decoded samples from real images $x \sim p_{\text{data}}$:
\begin{equation}
\mathcal{L}_{\text{adv}}^{g}(\phi)
=
\mathbb{E}_{z_g \sim p_g}
\Big[
\mathcal{L}_{\text{adv}}
\big(
\mathcal{D}_{\phi}(z_g);
\mathcal{C}_{\omega}
\big)
\Big].
\label{eq:lpgan_adv}
\end{equation}
We adopt the DINO-based discriminator and adversarial training setup of \citet{Tian2024VisualAM}, as used in prior tokenizer training works~\citep{Chen2024SoftVQVAEE1,Li2024ImageFolderAI}. Throughout adaptation, $\mathcal{G}_{\varphi}$ and $\mathcal{E}_{\theta}$ remain fixed; only $\mathcal{D}_{\phi}$ and $\mathcal{C}_{\omega}$ are updated. Consequently, LP-GANs leaves the generation-time latent distribution $p_g$ unchanged and confines adversarial optimization to the downstream decoder.

\begin{figure}[t]
    \centering
    \includegraphics[width=\linewidth]{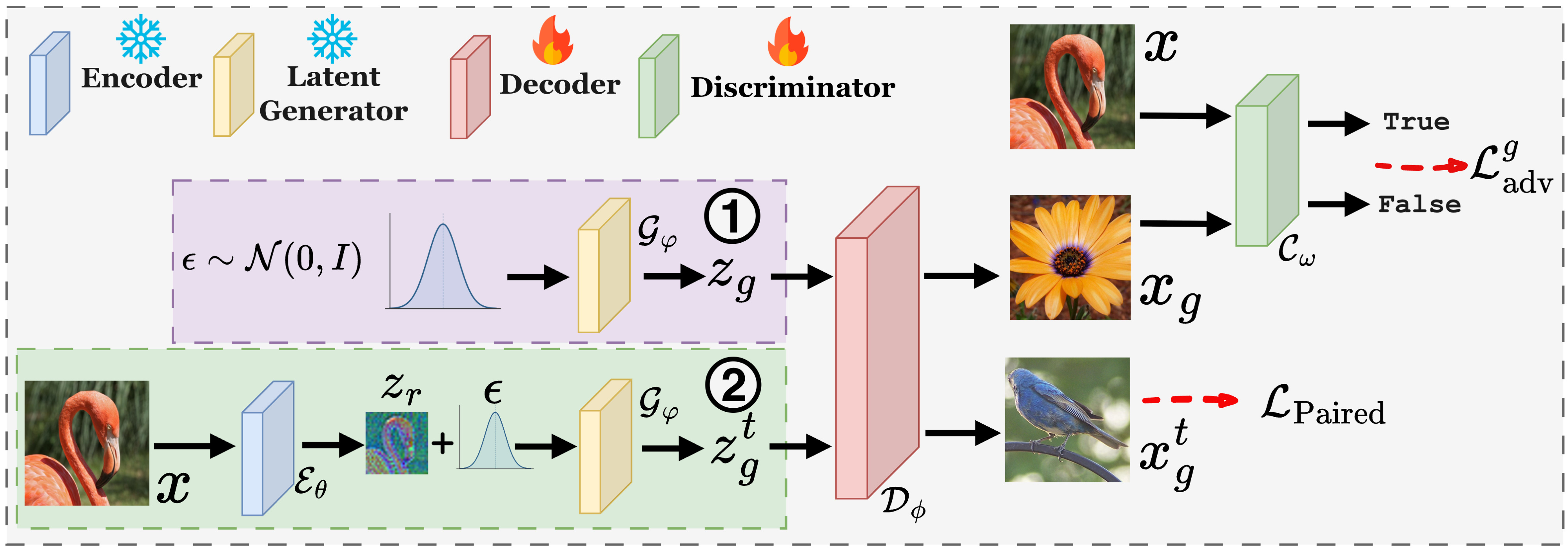}
\vspace{-12pt}
    \caption{\textbf{Overview of Latent Prior GANs (LP-GANs).}
     LP-GANs freeze the pretrained latent generator and refine the decoder using two complementary objectives.
    \textbf{(1)} Standard generation produces $z_g \sim p_g$, and the decoded samples $x_g$ are adversarially matched to real images through $\mathcal{L}_{\text{adv}}^{g}$.
    \textbf{(2)} Starting from a reconstruction latent $z_r$, we perturb and partially sample it with the frozen generator to obtain a generation-aware intermediate latent $z_g^t$ that retains correspondence with $x$, enabling paired supervision through $\mathcal{L}_{\text{pair}}$. Together, these objectives adapt the decoder to generation-time latents while anchoring its refinement to the inherited latent-to-image mapping.}
    \label{fig:method}
\end{figure}

\subsection{Generation-Aware Paired Supervision}
\label{sec:paired_supervision}
\Cref{eq:lpgan_adv} aligns the distribution of decoded samples with the data distribution but does not constrain their sample-wise correspondence to the input latents. Optimized in isolation, the decoder may therefore drift from its inherited latent-to-image mapping. Paired reconstruction and perceptual objectives can anchor this adaptation, but this requires source-image correspondence, which fully generated latents do not provide. We therefore construct generation-aware intermediate latents that move toward the generation-time distribution while remaining paired with their source images.

Given $x \sim p_{\text{data}}$, we first obtain $z_r = \mathcal{E}_{\theta}(x)$ and perturb it toward the sampler initialization as $z_{\text{noisy}}^{t} = a_t z_r + b_t \epsilon$, where $\epsilon \sim \mathcal{N}(0,I)$ and the noise level $\eta_t = b_t/(a_t+b_t)$ interpolates between $z_r$ at $\eta_t=0$ and the standard noise initialization at $\eta_t=1$. Starting from $z_{\text{noisy}}^{t}$, we run the corresponding partial sampling procedure to obtain
\begin{equation*}
z_g^{t}
=
\mathcal{S}_t
\big(
\mathcal{G}_{\varphi},
z_{\text{noisy}}^{t},
\xi
\big)
\sim p_g^{t}(z),
\end{equation*}
where $\mathcal{S}_t$ denotes the deployed sampler run from the corresponding noise level to the generation endpoint. As $\eta_t$ increases, $p_g^{t}$ moves toward the generation-time distribution. In particular, at $\eta_t=1$, $z_{\text{noisy}}^{t}=\epsilon$ and the procedure reduces to standard generation, so $p_g^{1}=p_g$~\citep{Fang2026GenerationAwareRecon}.
Because $z_g^{t}$ originates from the reconstruction latent of $x$, it retains image-level correspondence with $x$.

We therefore define the paired objective as
\begin{equation}
\mathcal{L}_{\text{pair}}(\phi)
=
\mathbb{E}
\Big[
\mathcal{L}_{\text{rec}}
\big(
x,\mathcal{D}_{\phi}(z_g^{t})
\big)
+
\lambda_{\text{perc}}\,
\mathcal{L}_{\text{perc}}
\big(
x,\mathcal{D}_{\phi}(z_g^{t})
\big)
\Big],
\label{eq:paired_objective}
\end{equation}
where the expectation is over $x \sim p_{\text{data}}$, $t$, $\epsilon$, and $\xi$, with $t$ sampled uniformly during adaptation. The full decoder objective is
\begin{equation}
\mathcal{L}_{\text{LP-GANs}}
=
\mathcal{L}_{\text{adv}}^{g}
+
\lambda_{\text{pair}}\,
\mathcal{L}_{\text{pair}}.
\label{eq:lpgan_full}
\end{equation}
The adversarial term refines the decoder directly on the deployment distribution, while the paired term anchors this refinement to the inherited latent-to-image mapping.

\section{Experiments}
\label{sec:experiments}


Our experiments evaluate LP-GANs across diffusion/flow-based and autoregressive latent generators. We first compare LP-GANs with existing diffusion, flow, GAN, and autoregressive models at the system level (Section~\ref{sec:system_comparison}). We then analyze LP-GANs on the iMF family~\citep{Geng2025ImprovedMF}, examining the contributions of adversarial refinement and paired supervision, sensitivity to the trajectory noise level, and the choice of components to refine (Section~\ref{sec:component_analysis}). Finally, we analyze the computational cost of decoder refinement (Section~\ref{sec:training_efficiency}) and evaluate its generalization beyond the iMF family (Section~\ref{sec:generalization}).

\paragraph{Experiment Setup.}
We conduct experiments on ImageNet-1K~\citep{russakovsky2015imagenet} at $256\times256$ resolution. Our primary evaluation uses the iMF family~\citep{Geng2025ImprovedMF}, including iMF-B/2, iMF-M/2, iMF-L/2, and iMF-XL/2, all sharing the same pretrained SD-VAE tokenizer~\citep{Rombach2022HighResolutionIS}. During LP-GANs training, the tokenizer encoder and latent generator remain frozen, while the decoder and a newly introduced discriminator are optimized. For generators supporting classifier-free guidance (CFG), decoder refinement uses generator-produced latents sampled under the corresponding CFG settings, together with generation-aware intermediate latents constructed using the same guidance configuration. Beyond the iMF family, we further evaluate LP-GANs on MeanFlow-XL/2~\citep{Geng2025MeanFF} and DMF-XL/2+~\citep{Lee2025DecoupledMT}. We additionally evaluate LP-GANs on TiTok-L32~\citep{Yu2024AnII} to investigate whether decoder refinement extends to autoregressive latent generation. Additional implementation details are provided in Appendix~\ref{app:training_details}.

\paragraph{Evaluation Metrics and Protocols.} We evaluate generative performance using Fr\'echet Inception Distance (gFID)~\citep{Heusel2017GANsTB}. The iMF~\citep{Geng2025ImprovedMF} and MeanFlow~\citep{Geng2025MeanFF} papers use a distinct FID evaluation protocol, which we refer to as the \emph{official iMF protocol}, alongside the OpenAI protocol used in our experiments. Unless otherwise specified, we report gFID under the OpenAI protocol and restrict comparisons to results obtained under the same protocol. Absolute gFID values can differ between protocols, even for the same model. We provide details of both protocols in Appendix~\ref{appendix:gfid_protocol}.


\subsection{System-Level Comparison}
\label{sec:system_comparison}

\begin{table}[!t]
\centering
\caption{\small{\textbf{System-level comparison on class-conditional ImageNet 256${\times}$256}. NFE denotes the number of function evaluations during inference. All numbers are with CFG when applicable. \textbf{Bold} indicates the best result in each comparison. $^{\dagger}$ denotes results from~\citep{Geng2025ImprovedMF}, $^{\ddagger}$ denotes results cited from GAR~\citep{Fang2026GenerationAwareRecon}, and $^{\star}$ denotes results reproduced or obtained by us under the official iMF protocol.
}} 
\vspace{-0.5ex}
\resizebox{0.32\linewidth}{!}{
\setlength{\tabcolsep}{4pt}
\small
\begin{tabular}{@{}lccc@{}}
\toprule
{Method} & {\# Params} & NFE & {FID} \\
\midrule
\rowcolor[gray]{0.9}
\multicolumn{4}{l}{\textit{\textbf{1-NFE diffusion/flow from scratch}}} \\
MeanFlow-XL/2$^{\star}$ & 676M & 1 & 3.34   \\
\textbf{MeanFlow-XL/2+LP-GANs (Ours)}$^{\star}$ & 676M & 1 &  2.94  \\
TiM-XL/2$^{\dagger}$ & 664M & 1 & 3.26 \\
$\alpha$-Flow-XL/2+$^{\dagger}$ & 676M & 1 & 2.58 \\
\midrule
iMF-B/2$^{\star}$ & 89M  & 1 & 3.37             \\
iMF-M/2$^{\star}$ & 174M & 1 & 2.27             \\
iMF-L/2$^{\star}$ & 409M & 1 & 1.86             \\
iMF-XL/2$^{\star}$ & 610M & 1 & 1.73    \\
\midrule
iMF-B/2 + GAR$^{\ddagger}$    & 89M  & 1 & 2.81          \\
iMF-M/2+ GAR$^{\ddagger}$     & 174M & 1 &  1.98         \\
iMF-L/2+ GAR$^{\ddagger}$      & 409M & 1 & 1.58           \\
iMF-XL/2 + GAR$^{\ddagger}$  & 610M & 1 &   1.50  \\
\midrule
\textbf{iMF-B/2 + LP-GANs (Ours)}$^{\star}$ & 89M & 1 & \textbf{2.08} \\
\textbf{iMF-M/2 + LP-GANs (Ours)}$^{\star}$ & 174M & 1 & \textbf{1.73} \\
\textbf{iMF-L/2 + LP-GANs (Ours)}$^{\star}$ & 409M & 1 & \textbf{1.48} \\
\textbf{iMF-XL/2 + LP-GANs (Ours)}$^{\star}$ & 610M & 1 & \textbf{1.32} \\
\midrule
\rowcolor[gray]{0.9}
\multicolumn{4}{l}{\textit{\textbf{2-NFE diffusion/flow from scratch}}} \\
IMM-XL/2$^{\dagger}$ & 675M & 1$\times$2 & 7.77 \\
\bottomrule
\end{tabular}
}
\hfill
\resizebox{0.32\linewidth}{!}{
\setlength{\tabcolsep}{4pt}
\small
\begin{tabular}{@{}lccc@{}}
\toprule
{Method} & {\# Params} & NFE & FID \\
\midrule
\rowcolor[gray]{0.9}
\multicolumn{4}{l}{\textit{\textbf{2-NFE diffusion/flow from scratch}}} \\
MeanFlow-XL/2+$^{\dagger}$ & 676M & 2 & 2.20  \\
$\alpha$-Flow-XL/2+$^{\dagger}$ & 676M & 2 & 1.95\\
iMF-XL/2$^{\dagger}$ & 610M & 2 & 1.54   \\
\midrule
\multicolumn{4}{l} {\textit{\textbf{1-NFE diffusion/flow (distillation)}}} \\
$\pi$-Flow-XL/2$^{\dagger}$ & 675M & 1 & 2.85\\
{DMF-XL/2+}$^{\star}$ & 675M &  1 & 2.22\\
\textbf{{DMF-XL/2+} + LP-GANs (Ours)}$^{\star}$& 675M & 1 &  1.91  \\
FACM-XL/2$^{\dagger}$ & 675M & 1 & 1.76 \\
\midrule
\multicolumn{4}{l} {\textit{\textbf{Multi-NFE diffusion/flow}}} \\
ADM-G$^{\dagger}$ & 554M & 250$\times$2 & 4.59 \\
LDM-4-G$^{\dagger}$ & 400M & 250$\times$2 & 3.60 \\
SimDiff$^{\dagger}$  & 2B & 1000$\times$2 & 2.77 \\
DiT-XL/2$^{\dagger}$ & 675M & 250$\times$2 & 2.27\\
SiT-XL/2$^{\dagger}$ & 675M   & 250$\times$2 & 2.06  \\
SiT-XL/2\;+\;REPA$^{\dagger}$ & 675M & 250$\times$2 & 1.42 \\
SiD2$^{\dagger}$ & -- & 512$\times$2 & 1.38 \\
LightningDiT-XL/2$^{\dagger}$ & 675M & 250$\times$2 & 1.35 \\
DDT-XL/2$^{\dagger}$ & 675M  & 250$\times$2 & 1.26 \\
\bottomrule
\end{tabular}
}
\hfill
\resizebox{0.32\linewidth}{!}{
\setlength{\tabcolsep}{4pt}
\small
\begin{tabular}{@{}lccc@{}}
\toprule
{Method} & {\# Params} & {NFE} & FID \\
\midrule
\multicolumn{4}{l} {\textit{\textbf{Multi-NFE diffusion/flow}}} \\
RAE\;+\;DiT$^\text{DH}$-XL$^{\dagger}$ & 839M & 250$\times$2 & 1.13 \\
\midrule
\multicolumn{4}{l}{\textit{\textbf{GANs}}} \\
BigGAN$^{\dagger}$  & 112M & 1 & 6.95 \\
GigaGAN$^{\dagger}$ & 569M & 1 & 3.45 \\
StyleGAN-XL$^{\dagger}$ & 166M & 1 & 2.30 \\
\midrule
\multicolumn{4}{l}{\textit{\textbf{autoregressive/masking}}} \\
JetFormer-L$^{\dagger}$ & 2.75B & 256$\times$2 & 6.64 \\
MaskGIT$^{\dagger}$ & 227M & 8 & 6.18 \\
TiTok-L32$^{\star}$ & 307M & 8$\times$2 & 2.65 \\
\textbf{TiTok-L32 + LP-GANs (Ours)}$^{\star}$ & 307M & 8$\times$2 & 1.90 \\
RQ-Transformer$^{\dagger}$ & 3.8B & 256$\times$2 & 3.80 \\
STARFlow$^{\dagger}$  & 1.4B & 1024$\times$2 & 2.40 \\
LLamaGen-3B$^{\dagger}$ & 3.1B & 256$\times$2 & 2.18 \\
VAR-$d30$$^{\dagger}$  & 2B   & 10$\times$2 & 1.92 \\
MAR-H$^{\dagger}$  & 943M & 256$\times$2 & 1.55 \\
RAR-XXL$^{\dagger}$ & 1.5B & 256$\times$2 & 1.48\\
xAR-H$^{\dagger}$ & 1.1B & 50$\times$2 & 1.24 \\
\bottomrule
\end{tabular}%
}
\label{tab:all_results}
\vspace{-2pt}
\end{table}

\paragraph{System-Level Comparison Baselines.} We present a system-level comparison on class-conditional ImageNet $256\times256$ across diffusion/flow, GAN, and autoregressive or masked generative models, with inference cost measured by the number of function evaluations (NFE). For diffusion and flow models trained from scratch, we include 1-NFE methods MeanFlow-XL/2~\citep{Geng2025MeanFF}, TiM-XL/2~\citep{Wang2025TransitionMR}, $\alpha$-Flow-XL/2+~\citep{Zhang2025AlphaFlowUA}, and iMF~\citep{Geng2025ImprovedMF} at multiple scales (B/2, M/2, L/2, XL/2), together with 2-NFE methods IMM-XL/2~\citep{Zhou2025InductiveMM}, MeanFlow-XL/2+~\citep{Geng2025MeanFF}, $\alpha$-Flow-XL/2+~\citep{Zhang2025AlphaFlowUA}, and iMF-XL/2. We further include distillation-based 1-NFE methods $\pi$-Flow-XL/2~\citep{Chen2025piFlowPF}, DMF-XL/2+~\citep{Lee2025DecoupledMT}, and FACM-XL/2~\citep{Peng2025FlowAnchoredCM}, as well as multi-NFE diffusion/flow models ADM-G~\citep{Dhariwal2021DiffusionMB}, LDM-4-G~\citep{Rombach2022HighResolutionIS}, SimDiff~\citep{Hoogeboom2023simpleDE}, DiT-XL/2~\citep{Peebles2022ScalableDM}, SiT-XL/2~\citep{Ma2024SiTEF}, SiT-XL/2+REPA~\citep{Yu2024RepresentationAF}, SiD2~\citep{Hoogeboom2024SimplerD}, LightningDiT-XL/2~\citep{Yao2025ReconstructionVG}, DDT-XL/2~\citep{Wang2026DDTDD}, and RAE+DiT$^{\text{DH}}$-XL~\citep{zheng2026diffusion}. For broader reference, we also compare with GAN-based methods BigGAN~\citep{Brock2018LargeSG}, GigaGAN~\citep{Kang2023ScalingUG}, and StyleGAN-XL~\citep{Sauer2022StyleGANXLSS}, and autoregressive or masked models JetFormer-L~\citep{Tschannen2024JetFormerAA}, 
MaskGIT~\citep{Chang2022MaskGITMG}, TiTok-L32~\citep{Yu2024AnII}, RQ-Transformer~\citep{Lee2022AutoregressiveIG}, STARFlow~\citep{Gu2025STARFlowSL}, LLamaGen-3B~\citep{Sun2024AutoregressiveMB}, VAR-$d30$~\citep{Tian2024VisualAM}, MAR-H~\citep{Li2024AutoregressiveIG}, RAR-XXL~\citep{Yu2024RandomizedAV}, and xAR-H~\citep{Ren2025BeyondNN}.


\paragraph{System-Level Generative Performance.}
We compare LP-GANs with existing class-conditional image generation methods. As shown in Table~\ref{tab:all_results}, LP-GANs consistently improves gFID across the evaluated latent generative models without increasing inference cost. For the iMF family, gFID decreases from 3.37 to 2.08 on B/2, 2.27 to 1.73 on M/2, 1.86 to 1.48 on L/2, and 1.73 to 1.32 on XL/2, while retaining one-step generation. On MeanFlow-XL/2 and the distillation-based DMF-XL/2+, gFID decreases from 3.34 to 2.94 and from 2.22 to 1.91, respectively, both at $\text{NFE}=1$. For autoregressive latent generation, LP-GANs reduces TiTok-L32 gFID from 2.65 to 1.90 while preserving its eight-step sampling procedure. Among the evaluated 1-NFE diffusion/flow models, iMF-XL/2 + LP-GANs achieves the lowest gFID of 1.32.

\paragraph{Qualitative Results.} We provide qualitative comparisons in Appendix~\ref{app:qualitative} (Figures~\ref{fig:qual-xl-baseline-volcano}--\ref{fig:qual-selected-titok}). Corresponding samples share identical generated latents, with the decoder being the only component that differs between the compared outputs. Decoder refinement improves local details and texture quality while reducing visible artifacts. The paired-supervision ablation further shows that adversarial refinement alone can alter semantic content or spatial structure, whereas paired supervision helps preserve the correspondence between generated latents and decoded images. Similar visual improvements are observed across the other evaluated latent generators.



\subsection{Analysis of LP-GANs on the iMF Family}
\label{sec:component_analysis}
We examine three aspects of LP-GANs using the iMF family~\citep{Geng2025ImprovedMF}. We first assess the contributions of adversarial refinement and generation-aware paired supervision across four model scales (B/2, M/2, L/2, and XL/2). We then study sensitivity to the trajectory noise level and compare decoder-only, generator-only, and joint refinement to assess the choice of trainable components.

\begin{table}[!t]
\centering
\caption{\small{\textbf{LP-GANs across iMF model scales.} We report gFID without and with CFG across iMF model scales, comparing the pretrained baseline, LP-GANs without paired supervision (PS), and full LP-GANs under both evaluation protocols; full LP-GANs consistently perform best.
$^{\ddagger}$: our reproduction using the official iMF protocol. $^{\star}$: our reproduction using the OpenAI protocol. See Appendix~\ref{appendix:gfid_protocol} for protocol details.}}
\vspace{-0.5ex}
\label{tab:scale_protocol_results}
\setlength{\tabcolsep}{4pt}
\small
\begin{tabular*}{\linewidth}{@{\extracolsep{\fill}}lcccccccc@{}}
\toprule
\multirow{2}{*}{Setting} & \multicolumn{4}{c}{gFID w/o CFG ({\color{green!60!black}\bfseries$\pmb{\downarrow}$})} & \multicolumn{4}{c}{gFID w/ CFG ({\color{green!60!black}\bfseries$\pmb{\downarrow}$})} \\
\cmidrule(lr){2-5} \cmidrule(lr){6-9}
& B/2 & M/2 & L/2 & XL/2 & B/2 & M/2 & L/2 & XL/2 \\
\midrule
Baseline$^{\ddagger}$ & 16.58 & 11.92 & 9.28 & 9.78 & 3.37 & 2.27 & 1.86 & 1.73 \\
LP-GANs w/o PS$^{\ddagger}$ & 19.43 & 15.58 & 11.92 & 11.96 & 2.55 & 2.07 & 1.78 & 1.69 \\
LP-GANs w/ PS$^{\ddagger}$ & \textbf{13.39} & \textbf{11.20} & \textbf{8.49} & \textbf{8.51} & \textbf{2.08} & \textbf{1.73} & \textbf{1.48} & \textbf{1.32} \\
\midrule
Baseline$^{\star}$ & 16.41 & 11.92 & 9.26 & 9.41 & 3.47 & 2.40 & 1.90 & 1.80 \\
LP-GANs w/o PS$^{\star}$ & 18.99 & 15.11 & 11.63 & 11.12 & 2.56 & 2.06 & 1.82 & 1.67 \\
LP-GANs w/ PS$^{\star}$ & \textbf{13.64} & \textbf{10.66} & \textbf{8.52} & \textbf{8.52} & \textbf{2.14} & \textbf{1.77} & \textbf{1.51} & \textbf{1.51} \\
\bottomrule
\end{tabular*}
\vspace{-1ex}
\end{table}

\paragraph{Effect of Adversarial Refinement and Paired Supervision.}  Table~\ref{tab:scale_protocol_results} compares the pretrained baseline, adversarial refinement without paired supervision (PS), and full LP-GANs across four iMF model scales. With CFG, adversarial refinement alone improves gFID over the pretrained baseline under both evaluation protocols. Adding generation-aware paired supervision further reduces gFID at every scale. Under the official iMF protocol, gFID decreases from 2.55 to 2.08 on B/2, 2.07 to 1.73 on M/2, 1.78 to 1.48 on L/2, and 1.69 to 1.32 on XL/2. Without CFG, adversarial refinement alone increases gFID relative to the pretrained baseline, whereas adding paired supervision improves upon the baseline across all four scales. These results support the role of paired supervision in anchoring decoder refinement to the inherited latent-to-image mapping, as formulated in Equation~\ref{eq:lpgan_full}.

\begin{wrapfigure}[14]{r}{0.48\textwidth}
\wraptopalign
\captionsetup{width=\linewidth,skip=4pt}
\centering
\includegraphics[
  width=\linewidth,
  trim=35pt 0pt 20pt 0pt,
  clip
]{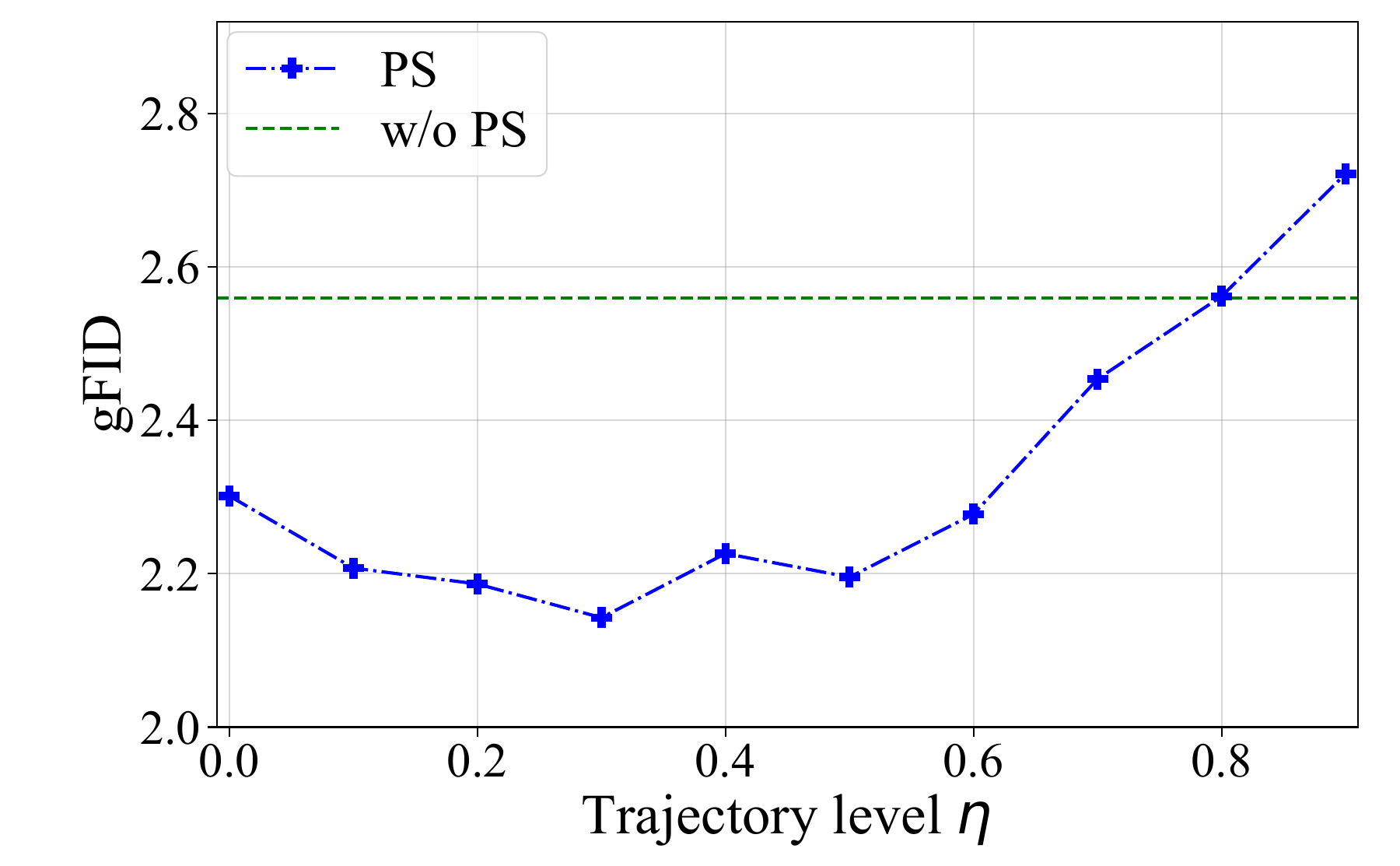}
\vspace{-12pt}
\caption{CFG gFID across trajectory level $\eta$.
The dashed line is LP-GANs w/o PS ($2.56$);
the curve is best at $\eta=0.3$.}
\label{fig:sensitivity}
\vspace{-4pt}
\end{wrapfigure}

\paragraph{Sensitivity to the Trajectory Noise Level.}
Generation-aware paired supervision depends on the trajectory noise level $\eta_t$, which controls the trade-off between proximity to generation-time latents and correspondence with the source image. We evaluate its effect on iMF-B/2. As shown in Figure~\ref{fig:sensitivity}, paired supervision improves gFID over adversarial refinement alone for $\eta_t \in [0,0.7]$, with the lowest gFID at $\eta_t=0.3$. Its benefit diminishes at higher noise levels, and performance deteriorates substantially as $\eta_t$ approaches 1. This trend is consistent with the reconstruction-to-generation trajectory studied in GAR~\citep{Fang2026GenerationAwareRecon}: increasing $\eta_t$ produces more generation-like latents but weakens their correspondence with the source image. At high noise levels, this correspondence may become too weak to provide reliable sample-wise reconstruction and perceptual supervision.



\begin{wraptable}[9]{r}{0.48\textwidth}
\wraptopalign
\captionsetup{width=\linewidth,skip=4pt}
\centering
\caption{\footnotesize{\textbf{Comparison of refinement scopes.} Results on iMF-B/2 under the OpenAI protocol. Decoder-only refinement achieves the lowest gFID with computational cost comparable to generator-only and joint refinement. Costs are measured in H100 GPU-hours per epoch.}}
\vspace{-1.0ex}
\label{tab:update_scope}
\footnotesize
\setlength{\tabcolsep}{3pt}
\begin{tabular*}{\linewidth}{@{\extracolsep{\fill}}lccc@{}}
\toprule
Scope & CFG & no-CFG & Cost \\
\midrule
Frozen & 3.47 & 16.41 & 0 \\
\textbf{Decoder-only} & \textbf{2.14} & \textbf{13.64} & 12.7 \\
Generator-only & 3.91 & 22.33 & \textbf{11.5} \\
Joint & 2.40 & 17.25 & 13.9 \\
\bottomrule
\end{tabular*}
\vspace{-4pt}
\end{wraptable}

\paragraph{Which Component Should Be Refined?}
We compare four update scopes on iMF-B/2: frozen, decoder-only, generator-only, and joint refinement. As shown in Table~\ref{tab:update_scope}, decoder-only refinement achieves the lowest gFID both with and without CFG. Under CFG, it reduces gFID from 3.47 to 2.14, outperforming joint refinement (2.40) and generator-only refinement (3.91). Generator-only refinement performs worse than the pretrained baseline under both evaluation settings. These results favor adapting the decoder to the pretrained generator's latent distribution rather than modifying the generator itself. The three trainable scopes have comparable costs: $12.7$, $11.5$, and $13.9$ H100 GPU-hours per epoch for decoder-only, generator-only, and joint refinement, respectively. Decoder-only refinement achieves the best generative performance among the evaluated update scopes with comparable computational cost.


\subsection{Training Efficiency}
\label{sec:training_efficiency}

We compare the cost of LP-GANs decoder refinement with the costs of training class-conditional GANs from scratch and pretraining the tokenizer and latent generator.

\vspace{-0.8em}
\Needspace{16\baselineskip}
\paragraph{Efficiency Compared with From-Scratch GAN Training.} 

\begin{wraptable}[9]{r}{0.52\textwidth}
\wraptopalign
\captionsetup{width=\linewidth,skip=4pt}
\centering
\caption{\footnotesize{\textbf{From-scratch GAN training versus LP-GANs decoder refinement.} Class-conditional ImageNet at $256 \times 256$ resolution. BigGAN and StyleGAN-XL costs are FLOPs-based H100-equivalent estimates, whereas LP-GANs cost is measured on H100 GPUs and covers only decoder refinement, excluding tokenizer and latent-generator pretraining.}}
\vspace{-0.8ex}
\label{tab:gan_cost}
\footnotesize
\setlength{\tabcolsep}{3pt}
\begin{tabular*}{\linewidth}{@{\extracolsep{\fill}}lcc@{}}
\toprule
Method & FID $\downarrow$ & H100-h $\downarrow$ \\
\midrule
BigGAN\citep{Brock2018LargeSG} & 6.95 & 5{,}000 \\
StyleGAN-XL\citep{Sauer2022StyleGANXLSS} & 2.30 & 2{,}000 \\
\textbf{LP-GANs (iMF-XL/2)} & \textbf{1.32} & \textbf{88} \\
\bottomrule
\end{tabular*}
\vspace{-1pt}
\end{wraptable}

To better understand the training efficiency of LP-GANs, we compare its training cost and generative performance with those of BigGAN and StyleGAN-XL on ImageNet (Table~\ref{tab:gan_cost}). Conventional GANs rely on adversarial optimization to learn generative structure from scratch, requiring substantially more adversarial training. LP-GANs instead inherits this structure from a pretrained latent generative model and restricts adversarial optimization to decoder refinement. For iMF-XL/2, five epochs of decoder refinement require 88 H100 GPU-hours and achieve a gFID of 1.32, compared with estimated training costs of 2{,}000 to 5{,}000 H100-equivalent GPU-hours for the from-scratch GAN baselines. Although these costs are not directly comparable as end-to-end training budgets, since LP-GANs excludes tokenizer and latent-generator pretraining, the comparison illustrates the relatively low additional cost of adversarial refinement when a pretrained latent generative model is available.


\paragraph{Refinement Cost Relative to Latent-Prior Training.}

\begin{wraptable}[9]{r}{0.48\textwidth}
\wraptopalign
\captionsetup{width=\linewidth,skip=4pt}
\centering
\caption{\footnotesize{\textbf{Latent-prior pretraining vs.\ LP-GANs refinement.}
Costs are estimated total wall-clock GPU-hours on a single H100 GPU; tokenizer and latent-generator costs are FLOPs-based estimates.}}
\vspace{-0.5ex}
\label{tab:priorcost}
\footnotesize
\setlength{\tabcolsep}{3pt}
\begin{tabular*}{\linewidth}{@{\extracolsep{\fill}}lccc@{}}
\toprule
Model & Tokenizer & Latent Generator & LP-GANs \\
\midrule
iMF-B/2  & $750$ & $440$  & $\mathbf{64}$ \\
iMF-M/2  & $750$ & $860$  & $\mathbf{66}$ \\
iMF-L/2  & $750$ & $2{,}000$ & $\mathbf{72}$ \\
iMF-XL/2 & $750$ & $3{,}800$ & $\mathbf{88}$ \\
\bottomrule
\end{tabular*}
\vspace{-4pt}
\end{wraptable}
Table~\ref{tab:priorcost} compares LP-GANs refinement costs with the estimated pretraining costs of the SD-VAE tokenizer and iMF latent generators across four model scales. The tokenizer is shared across all four models, while latent-generator pretraining costs increase from 440 H100-equivalent GPU-hours for iMF-B/2 to 3,800 for iMF-XL/2. In comparison, five epochs of decoder refinement require 64-88 measured H100 GPU-hours across these scales. LP-GANs therefore adds a relatively small computational overhead to the pretrained generative systems without requiring further training of the tokenizer or latent generator. 


\subsection{Generalization to Other Latent Generators}
\label{sec:generalization}

\begin{wraptable}[9]{r}{0.43\textwidth}
\wraptopalign
\captionsetup{width=\linewidth,skip=4pt}
\centering
\caption{\footnotesize{\textbf{Generalization to MeanFlow-XL/2 and DMF-XL/2+.} gFID under the OpenAI and official iMF evaluation protocols; lower is better. Both models use their respective pretrained sampling settings without additional CFG.}}
\vspace{-1ex}
\label{tab:generalization}
\footnotesize
\setlength{\tabcolsep}{3pt}
\begin{tabular*}{\linewidth}{@{\extracolsep{\fill}}llcc@{}}
\toprule
Generator & Protocol & Baseline & LP-GANs \\
\midrule
\multirow{2}{*}{MeanFlow-XL/2} & OpenAI & 3.49 & \textbf{2.98} \\
& Official & 3.34 & \textbf{2.94} \\
\multirow{2}{*}{DMF-XL/2+} & OpenAI & 2.40 & \textbf{1.95} \\
& Official & 2.22 & \textbf{1.91} \\
\bottomrule
\end{tabular*}
\vspace{-4pt}
\end{wraptable}

\paragraph{Generalization Beyond the iMF Family.}
We further evaluate LP-GANs on MeanFlow-XL/2~\citep{Geng2025MeanFF} and DMF-XL/2+~\citep{Lee2025DecoupledMT}. As shown in Table~\ref{tab:generalization}, LP-GANs improves gFID over the pretrained baselines under both evaluation protocols. For MeanFlow-XL/2, gFID decreases from $3.49$ to $2.98$ under the OpenAI protocol and from $3.34$ to $2.94$ under the official iMF protocol. For DMF-XL/2+, the corresponding improvements are from $2.40$ to $1.95$ and from $2.22$ to $1.91$. These results extend the improvements observed on the iMF family to two additional diffusion- and flow-based latent generators.



\Needspace{14\baselineskip}
\begin{wrapfigure}[10]{r}{0.49\textwidth}
\wraptopalign
\vspace{-2pt}
\captionsetup{width=\linewidth,skip=4pt}
\centering
\includegraphics[width=\linewidth]{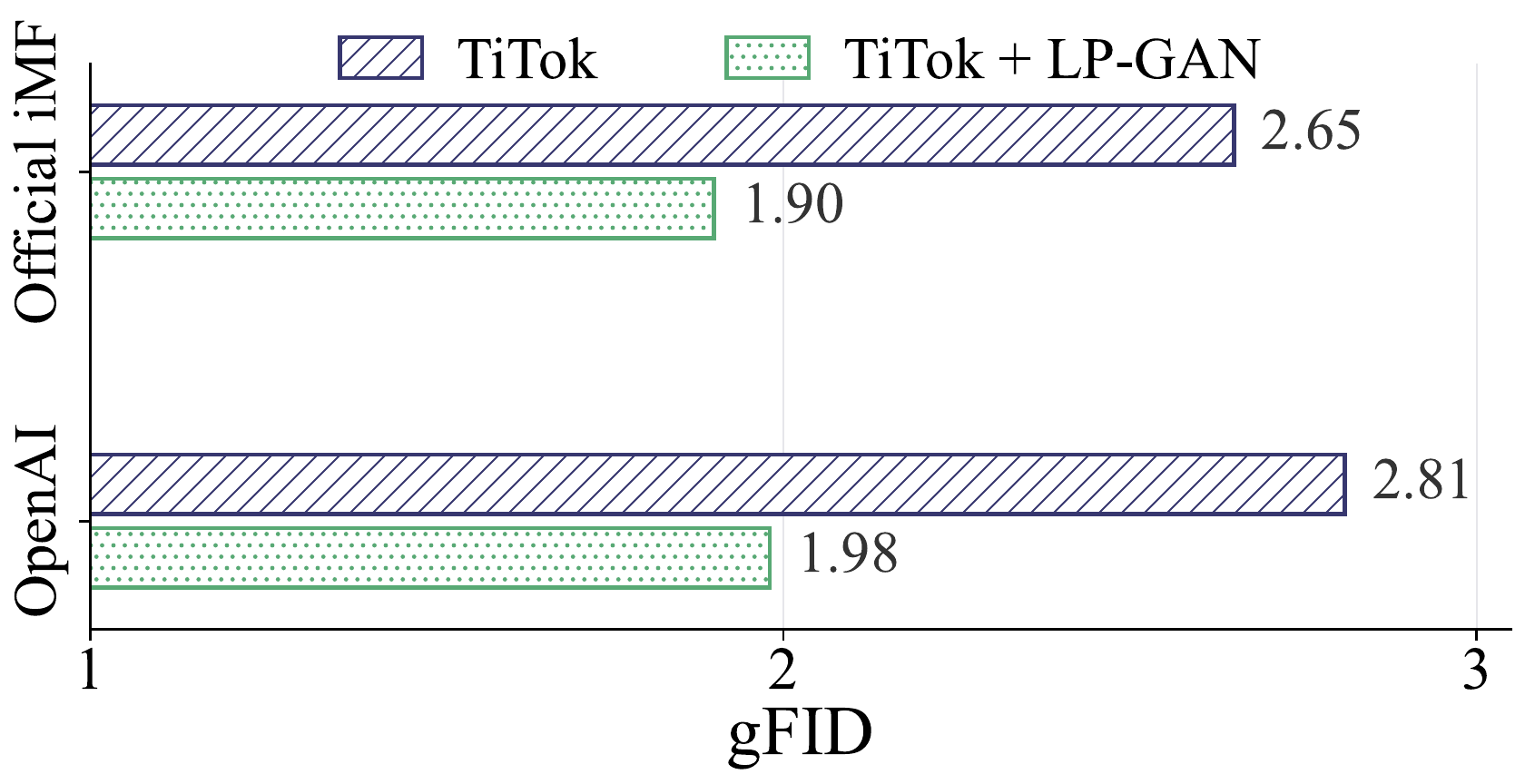}
\vspace{-15pt}
\caption{TiTok-L32 gFID under both protocols.}
\label{fig:titok_gfid}
\vspace{-4pt}
\end{wrapfigure}

\paragraph{Generalization to Autoregressive Generators.}
We also evaluate LP-GANs on autoregressive latent generation. Unlike diffusion- and flow-based generators, autoregressive generators do not naturally provide the reconstruction-to-generation trajectory used to construct generation-aware intermediate latents. We therefore retain adversarial refinement on fully generated latents but use encoder-induced reconstruction latents $z_r=\mathcal{E}_\theta(x)$ for paired reconstruction and perceptual supervision. Specifically, we refine the decoder of TiTok-L32~\citep{Yu2024AnII} while keeping its pretrained MaskGIT generator and eight-step sampling procedure unchanged. This allows LP-GANs to adapt the decoder to generation-time latents while maintaining paired supervision without relying on a continuous generation trajectory.

As shown in Figure~\ref{fig:titok_gfid}, LP-GANs improves TiTok-L32 under both evaluation protocols. Under the OpenAI protocol, gFID decreases from 2.81 to 1.98, while under the official iMF protocol it decreases from 2.65 to 1.90. With the pretrained encoder and latent generator frozen, these improvements are obtained through decoder refinement alone.

\section{Conclusion}
\label{sec:conclusion}

We introduced \textbf{Latent Prior GANs (LP-GANs)}, a framework for efficient decoder refinement via inherited generative structure. With a strong latent generative model available, adversarial learning no longer has to build generative structure from scratch. It only improves generation under the inherited latent prior. The decoder is then the part that needs to change: it was trained on reconstruction latents, but at generation time it receives generated latents. LP-GANs directly address this decoder train-generation mismatch by adapting the decoder to generation-time latents, anchored by generation-aware paired supervision. We observe the same result for every generator and scale we test: five epochs of adversarial adaptation already give better samples than the pretrained model. We hope to extend this approach to additional generative architectures, datasets, and generation settings in future work.

\printbibliography

\clearpage
\appendix
\section{Implementation and Training Details}
\label{app:training_details}
\label{appendix:experimental details}

\paragraph{Data and preprocessing.} All experiments use ImageNet-1K~\citep{russakovsky2015imagenet} at $256\times256$ resolution. Images are preprocessed using the iterative box-downsampling procedure adopted in LlamaGen~\citep{Sun2024AutoregressiveMB}, followed by random cropping to $256\times256$ and random horizontal flipping during training.

\paragraph{Training setup.} For the iMF family~\citep{Geng2025ImprovedMF}, MeanFlow-XL/2~\citep{Geng2025MeanFF}, and DMF-XL/2+~\citep{Lee2025DecoupledMT}, we freeze the pretrained VAE encoder and latent generator and optimize only the decoder and a DINO-based image discriminator. Both generator-produced latents and generation-aware intermediate latents are constructed under the corresponding guidance settings. All models are trained for five epochs on eight NVIDIA H100 GPUs with a batch size of $8$ per GPU (global batch size $64$). The decoder and discriminator use separate AdamW optimizers~\citep{loshchilov2017decoupled}, each with $(\beta_1,\beta_2)=(0.9,0.95)$ and weight decay $10^{-4}$. Both learning rates are initialized to $10^{-5}$. The decoder learning rate is halved after each epoch, with a minimum of $10^{-6}$, while the discriminator learning rate remains fixed at $10^{-5}$. Each training iteration consists of one decoder update followed by one discriminator update. Gradients are clipped to a maximum norm of $1.0$ for both optimizers.

\paragraph{Losses.} We use hinge adversarial losses~\citep{lim2017geometric} for adversarial training. The paired objective combines reconstruction loss with a perceptual loss. The perceptual loss consists of VGG-based LPIPS~\citep{Zhang2018TheUE} and a feature-space MSE computed using an ImageNet-pretrained ConvNeXt-S~\citep{liu2022convnet}, weighted by $1.0$ and $0.1$, respectively. The adversarial loss is adaptively weighted by $0.7$ times the ratio of the paired-loss gradient norm to the adversarial-loss gradient norm, computed with respect to the decoder's final layer~\citep{Esser2021TamingTF}.

\paragraph{Discriminator.}
We adopt a DINO-pretrained ViT-S/16~\citep{caron2021emerging} as the image discriminator, following the VAR setup~\citep{Tian2024VisualAM}. Discriminator training incorporates LeCAM regularization~\citep{tseng2021regularizing} with weight $0.001$, consistency regularization~\citep{zhang2019consistency} with weight $4.0$, and differentiable DiffAugment~\citep{zhao2020differentiable} using color, translation, and cutout transformations with strength $0.5$.

\paragraph{Noise Level Selection.} The intermediate noise level $\eta$ controls the construction of generation-aware intermediate latents used for paired supervision. The optimal noise level varies across pretrained latent generators, although we generally observe favorable performance for $\eta$ between $0.2$ and $0.4$. Specifically, we use $\eta=0.3$ for iMF-B/2, $\eta=0.2$ for iMF-M/2, iMF-L/2, iMF-XL/2, and MeanFlow-XL/2, and $\eta=0.4$ for DMF-XL/2+.

\paragraph{Autoregressive Generation.} Unlike diffusion and flow models, TiTok-L32~\citep{Yu2024AnII} uses discrete latent tokens generated by a pretrained MaskGIT model~\citep{Chang2022MaskGITMG}, without a continuous denoising trajectory for constructing generation-aware intermediate latents. We therefore use generator-produced tokens for adversarial supervision and encoder-induced tokens for paired supervision. The pretrained generator remains frozen and uses eight sampling steps with a CFG scale of $4.5$. Training follows the same five-epoch setup on eight NVIDIA H100 GPUs, but with a batch size of $2$ per GPU, zero weight decay, and both learning rates halved after each epoch to a minimum of $10^{-6}$. We use the same gradient-norm-based adaptive adversarial weighting as for the diffusion and flow models, but with a scaling factor of $1.0$.

\paragraph{Compute accounting.} All LP-GANs refinement costs are measured from actual H100 GPU runs. For iMF-XL/2, decoder-only refinement requires $17.6$ H100 GPU-hours per epoch, totaling $88$ H100 GPU-hours over five epochs. The per-epoch costs for the update-scope ablation on iMF-B/2 (Table~\ref{tab:update_scope}) are $12.7$, $11.5$, and $13.9$ H100 GPU-hours for decoder-only, generator-only, and joint refinement, respectively. These variants use the same training configuration and differ only in their trainable parameter sets. In contrast, the training costs of BigGAN and StyleGAN-XL in Table~\ref{tab:gan_cost} are based on FLOPs estimates or values reported in the original papers, rather than direct H100 measurements. Since their original training was conducted on Google TPUs and NVIDIA V100 GPUs, respectively, these estimates are not directly comparable to our measured H100 GPU-hours.

\Needspace{18\baselineskip}
\begin{wraptable}{r}{0.50\textwidth}
\wraptopalign
\captionsetup{width=\linewidth,skip=4pt}
\centering
\vspace{-0.5ex}
\caption{\footnotesize{\textbf{Training CFG settings.} Guidance used to construct latents for decoder refinement; -- denotes a non-applicable interval.}}
\label{tab:cfg_training}
\vspace{-0.0ex}
\footnotesize
\setlength{\tabcolsep}{3pt}
\begin{tabular*}{\linewidth}{@{\extracolsep{\fill}}lccc@{}}
\toprule
Generator & $\omega$ & $t_{\min}$ & $t_{\max}$ \\
\midrule
iMF-B/2  & $8.0$  & $0.40$ & $0.65$ \\
iMF-M/2  & $10.5$ & $0.40$ & $0.60$ \\
iMF-L/2  & $10.5$ & $0.40$ & $0.60$ \\
iMF-XL/2 & $8.0$  & $0.42$ & $0.62$ \\
MeanFlow-XL/2 & $1.0$ & $0.00$ & $1.00$ \\
DMF-XL/2+ & $1.0$ & $0.00$ & $1.00$ \\
TiTok-L32 & $4.5$ & -- & -- \\
\bottomrule
\end{tabular*}
\vspace{-4pt}
\end{wraptable}
\paragraph{Training CFG.} During decoder refinement, iMF uses the pretrained generators' default CFG settings for both generator-produced and generation-aware intermediate latents (Table~\ref{tab:cfg_training}). TiTok-L32 uses its native guidance scale with a linear eight-step schedule; continuous-time intervals do not apply. MeanFlow-XL/2 incorporates guidance into its pretrained generator, whereas DMF-XL/2+ uses an unguided checkpoint. Both run without additional sampling-time CFG ($\omega=1$).

\Needspace{17\baselineskip}
\begin{wraptable}{r}{0.50\textwidth}
\wraptopalign
\captionsetup{width=\linewidth,skip=4pt}
\centering
\vspace{-0.0ex}
\caption{\footnotesize{\textbf{Inference CFG settings.} Guidance used for quantitative evaluation of the refined decoders.}}
\label{tab:cfg_settings}
\vspace{-0.5ex}
\footnotesize
\setlength{\tabcolsep}{3pt}
\begin{tabular*}{\linewidth}{@{\extracolsep{\fill}}lccc@{}}
\toprule
Generator & $\omega$ & $t_{\min}$ & $t_{\max}$ \\
\midrule
iMF-B/2  & $8.0$  & $0.40$ & $0.65$ \\
iMF-M/2  & $10.5$ & $0.40$ & $0.60$ \\
iMF-L/2  & $8.0$  & $0.44$ & $0.60$ \\
iMF-XL/2 & $5.25$ & $0.48$ & $0.65$ \\
MeanFlow-XL/2 & $1.0$ & $0.00$ & $1.00$ \\
DMF-XL/2+ & $1.0$ & $0.00$ & $1.00$ \\
TiTok-L32 & $4.5$ & -- & -- \\
\bottomrule
\end{tabular*}
\vspace{-4pt}
\end{wraptable}
\paragraph{Inference CFG.} Decoder refinement can change which CFG settings yield the lowest gFID. We therefore search nearby guidance scales and intervals for iMF-L/2 and iMF-XL/2 after refinement (Table~\ref{tab:cfg_settings}). The remaining models retain their training-time settings. These inference settings are used to evaluate the refined decoders; pretrained baselines use the generators' default settings.

\section{Limitations}
Our experiments are limited to class-conditional ImageNet generation at $256\times256$ resolution. We have not evaluated LP-GANs on text-to-image generation, higher resolutions, or other datasets. In addition, the intermediate noise level $\eta$ requires model-specific tuning. Although values between $0.2$ and $0.4$ generally yield favorable results, the optimal value varies across latent generators (Table~\ref{tab:eta_sweep}), and we currently select it through a parameter sweep. Finally, our approach relies on a sufficiently capable pretrained latent generator, and its effectiveness may be limited when the underlying generator produces low-quality latent representations.

\clearpage
\section{gFID Evaluation Protocol}
\label{appendix:gfid_protocol}
\begin{table}[!htbp]
\centering
\caption{\small{Evaluation protocols on the iMF-B/2 baseline. The official iMF protocol uses the torch-fidelity pipeline and \texttt{jit\_in256\_stats.npz}; the OpenAI
protocol uses TensorFlow Inception and \texttt{VIRTUAL\_imagenet256\_labeled.npz}.}}
\vspace{-0.5ex}
\label{tab:fid_protocol_sensitivity}
\small
\resizebox{\linewidth}{!}{
\begin{tabular}{@{}lccc@{}}
\toprule
Protocol & Feature extractor & Reference statistics & gFID \\
\midrule
Official iMF & torch-fidelity & \texttt{jit\_in256\_stats.npz} & 3.39 \\
Official iMF (reproduced) & torch-fidelity & \texttt{jit\_in256\_stats.npz} & 3.37 \\
OpenAI & TensorFlow Inception & \texttt{VIRTUAL\_imagenet256\_labeled.npz} & 3.47 \\
\bottomrule
\end{tabular}
}
\vspace{-4pt}
\end{table}

The official iMF-B/2 evaluation~\citep{Geng2025ImprovedMF} differs from the commonly used OpenAI protocol in three aspects: (i) the computational platform (TPU vs.\ GPU), (ii) the Inception feature implementation and weights (the PyTorch torch-fidelity pipeline vs.\ TensorFlow Inception), and (iii) the reference real-image statistics (\texttt{jit\_in256\_stats.npz} vs.\ \texttt{VIRTUAL\_imagenet256\_labeled.npz}).
Absolute gFID values are not directly comparable across protocols, so all comparisons are made within the same protocol and reference statistics.

\section{Additional Trajectory-Level Sensitivity}
\label{app:Sensitivity}
Figure~\ref{fig:sensitivity} and Table~\ref{tab:eta_sweep} examine the effect of the intermediate noise level $\eta$ on iMF-B/2. Adversarial refinement without paired supervision achieves a gFID of $2.56$, serving as the reference. Introducing paired supervision improves gFID to $2.30$ even at $\eta=0$, with the best performance of $2.14$ achieved at $\eta=0.3$. The improvement persists up to $\eta=0.7$ but diminishes at higher noise levels, with gFID increasing to $2.72$ at $\eta=0.9$ and $8.67$ at $\eta=1.0$. This trend suggests that increasing $\eta$ progressively weakens the correspondence between generation-aware intermediate latents and their source images, eventually making paired supervision less reliable~\citep{Fang2026GenerationAwareRecon}.

\begin{table}[!h]
\centering
\caption{\small{\textbf{Trajectory-level sensitivity on iMF-B/2.}
All values are CFG gFID ($\downarrow$) under the OpenAI protocol.}}
\vspace{-0.5ex}
\label{tab:eta_sweep}
\small
\resizebox{\linewidth}{!}{
\begin{tabular}{@{}lrrrrrrrrrrr@{}}
\toprule
Trajectory level $\eta$ & 0.0 & 0.1 & 0.2 & 0.3 & 0.4 & 0.5 & 0.6 & 0.7 & 0.8 & 0.9 & 1.0 \\
\midrule
CFG gFID $\downarrow$ & 2.30 & 2.21 & 2.19 & \textbf{2.14} & 2.23 & 2.20 & 2.28 & 2.45 & 2.56 & 2.72 & 8.67 \\
\bottomrule
\end{tabular}
}
\vspace{-4pt}
\end{table}

\clearpage
\section{Qualitative Comparisons}
\label{app:qualitative}

We provide qualitative comparisons for iMF-XL/2, including the baseline comparison and paired-supervision ablation, as well as additional results for iMF-M/2, MeanFlow-XL/2, DMF-XL/2+, and TiTok-L32. Each figure contains two rows of eight samples. Unless otherwise specified, the top row shows outputs from the pretrained decoder, while the bottom row shows outputs from the LP-GANs-refined decoder. Corresponding samples share identical generator-produced latents, with the decoder being the only component that differs between the two rows. The paired-supervision ablation instead compares decoder refinement without and with paired supervision.

\subsection{iMF-XL/2: Baseline versus LP-GANs}

\begin{figure}[!htbp]
\centering
\includegraphics[width=\linewidth]{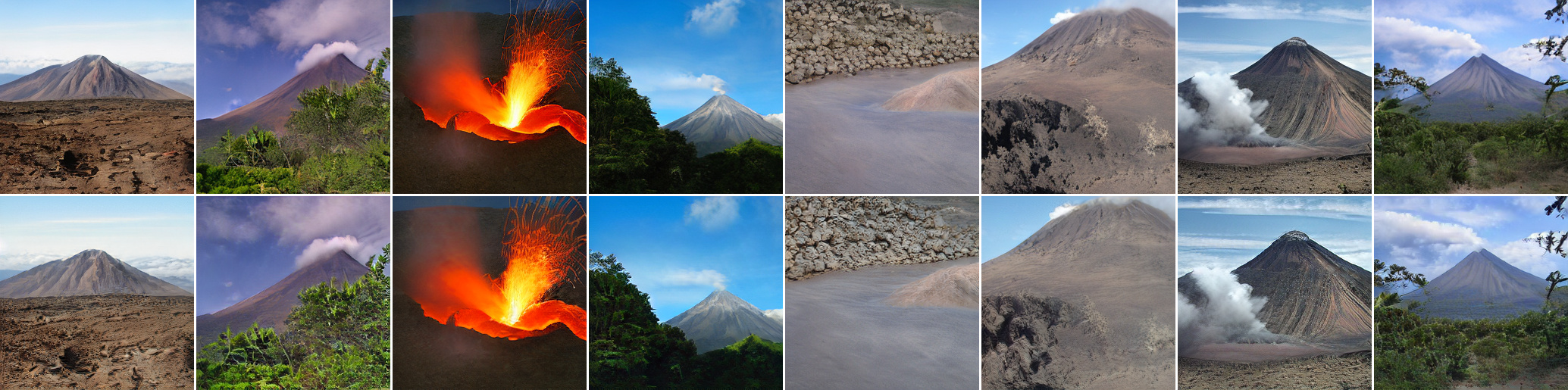}
\vspace{-3ex}
\caption{Class 980: volcano (top: before, bottom: after).}
\label{fig:qual-xl-baseline-volcano}
\vspace{-4pt}
\end{figure}

\begin{figure}[!htbp]
\centering
\includegraphics[width=\linewidth]{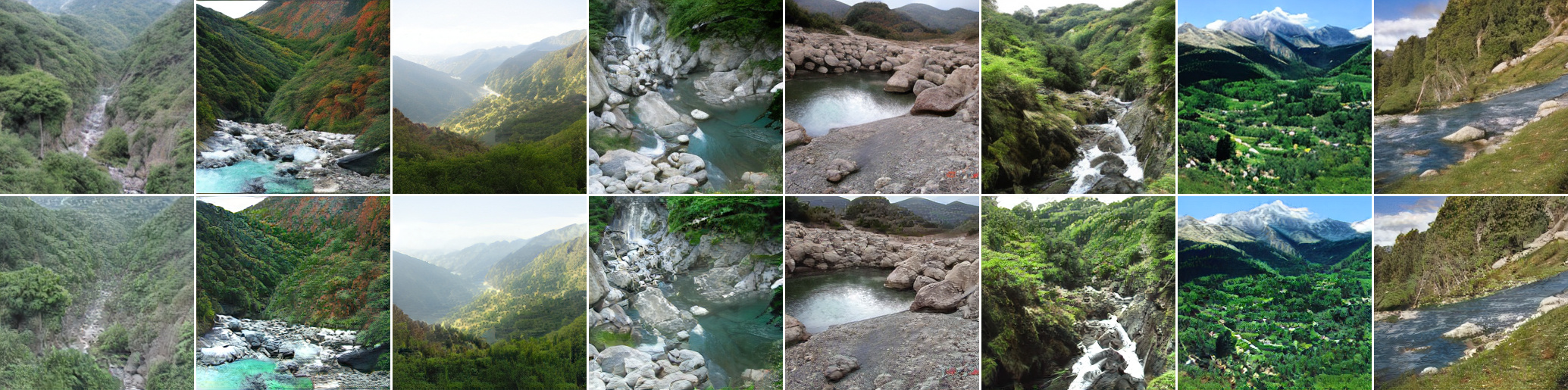}
\vspace{-3ex}
\caption{Class 979: valley (top: before, bottom: after).}
\label{fig:qual-xl-baseline-valley}
\vspace{-4pt}
\end{figure}

\begin{figure}[!htbp]
\centering
\includegraphics[width=\linewidth]{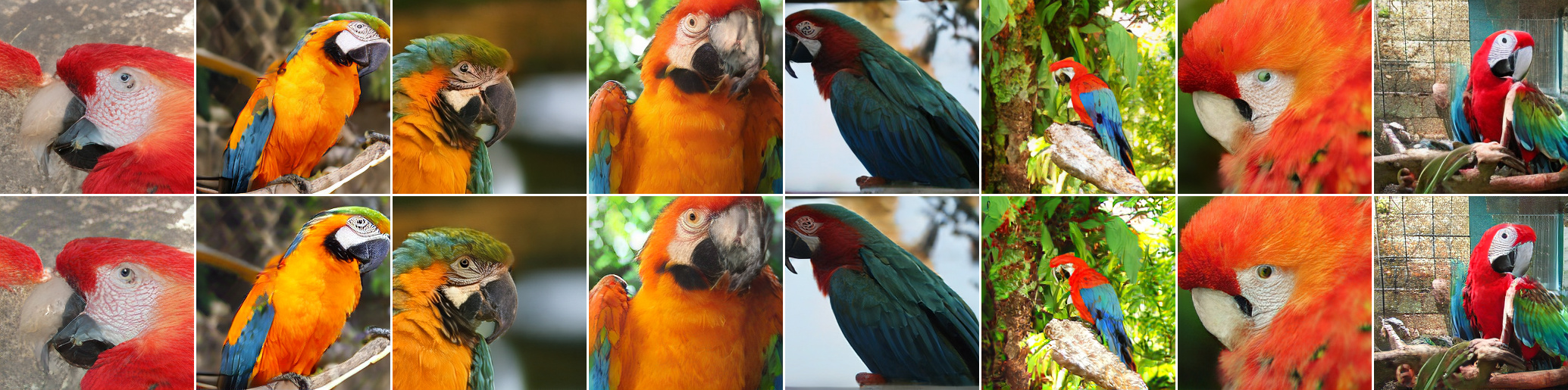}
\vspace{-3ex}
\caption{Class 88: macaw (top: before, bottom: after).}
\label{fig:qual-xl-baseline-macaw}
\vspace{-4pt}
\end{figure}

Figures~\ref{fig:qual-xl-baseline-volcano}--\ref{fig:qual-xl-baseline-macaw} compare the pretrained iMF-XL/2 decoder with its LP-GANs-refined counterpart across three ImageNet classes. The selected examples illustrate improvements in local detail and texture quality, with fewer visible artifacts while largely preserving the overall scene structure and class identity.

\clearpage
\subsection{iMF-XL/2: Effect of Paired Supervision}

\begin{figure}[!htbp]
\centering
\includegraphics[width=\linewidth]{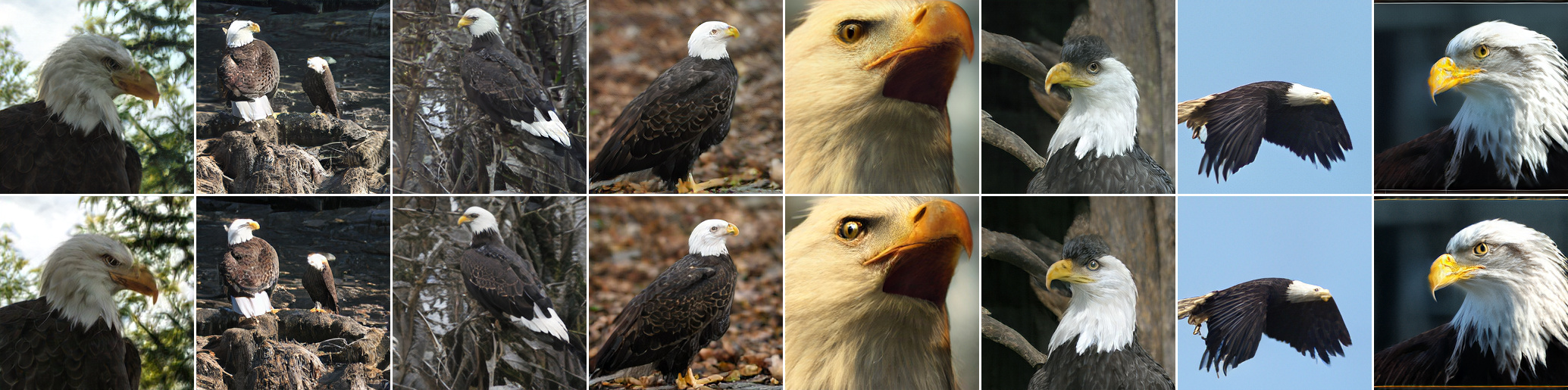}
\vspace{-3ex}
\caption{Class 22: bald eagle (top: without paired supervision, bottom: with paired supervision).}
\label{fig:qual-xl-paired-eagle}
\vspace{-4pt}
\end{figure}

\begin{figure}[!htbp]
\centering
\includegraphics[width=\linewidth]{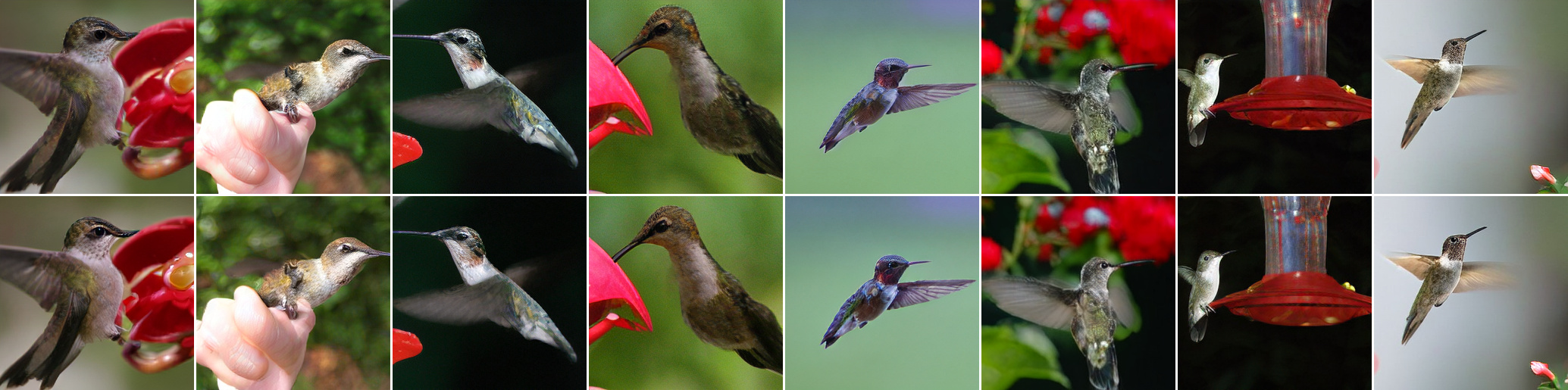}
\vspace{-3ex}
\caption{Class 94: hummingbird (top: without paired supervision, bottom: with paired supervision).}
\label{fig:qual-xl-paired-hummingbird}
\vspace{-4pt}
\end{figure}

\begin{figure}[!htbp]
\centering
\includegraphics[width=\linewidth]{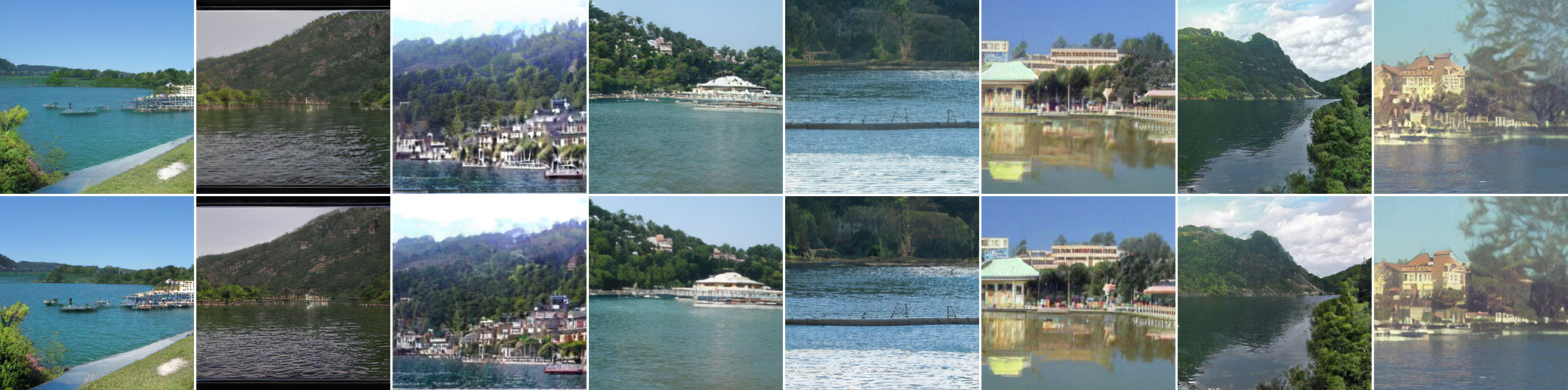}
\vspace{-3ex}
\caption{Class 975: lakeside (top: without paired supervision, bottom: with paired supervision).}
\label{fig:qual-xl-paired-lakeside}
\vspace{-4pt}
\end{figure}

Figures~\ref{fig:qual-xl-paired-eagle}--\ref{fig:qual-xl-paired-lakeside} illustrate the effect of paired supervision by comparing adversarial-only refinement (top) with the full LP-GANs objective (bottom). While adversarial-only refinement improves local visual quality, some samples exhibit changes in semantic content or spatial structure. Incorporating paired supervision helps preserve the correspondence between generated latents and their decoded images while retaining the visual improvements introduced by adversarial refinement.

\clearpage
\subsection{Other Latent Generators}

\begin{figure}[!htbp]
\centering
\includegraphics[width=\linewidth]{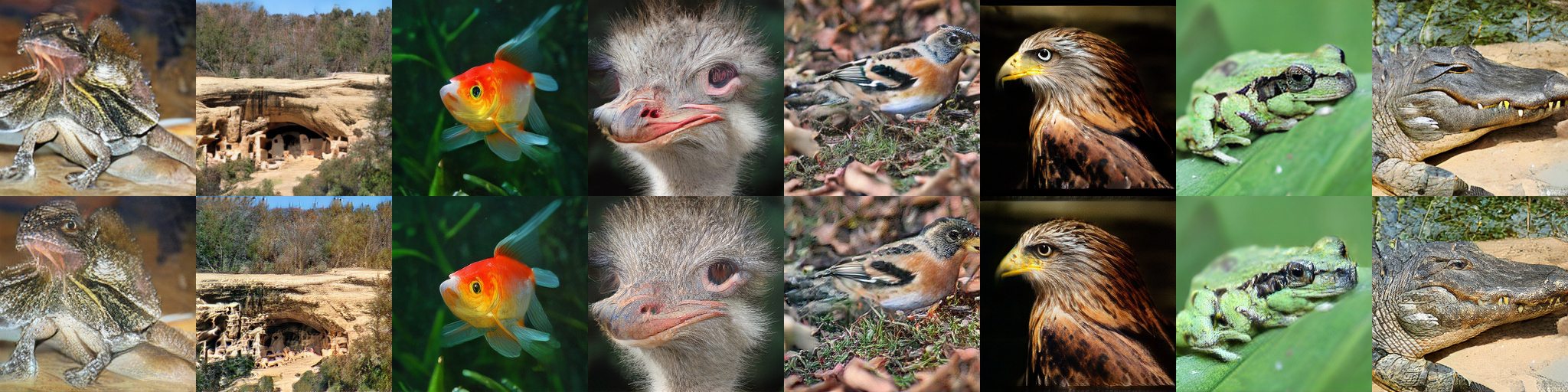}
\vspace{-3ex}
\caption{iMF-M/2: eight selected classes (top: before, bottom: after).}
\label{fig:qual-selected-imf-m}
\vspace{-4pt}
\end{figure}

\begin{figure}[!htbp]
\centering
\includegraphics[width=\linewidth]{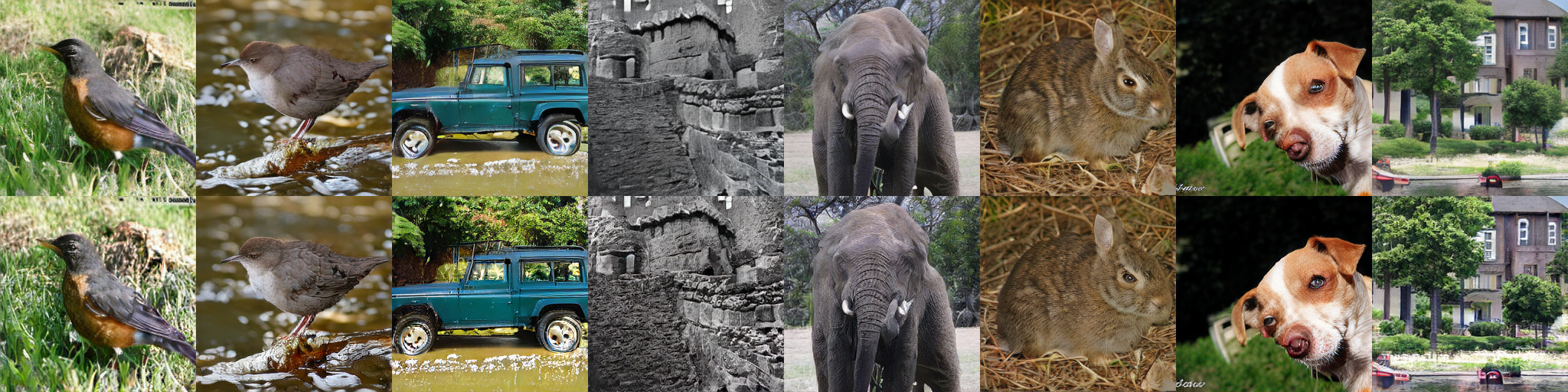}
\vspace{-3ex}
\caption{DMF-XL/2+: eight selected classes (top: before, bottom: after).}
\label{fig:qual-selected-dmf}
\vspace{-4pt}
\end{figure}

\begin{figure}[!htbp]
\centering
\includegraphics[width=\linewidth]{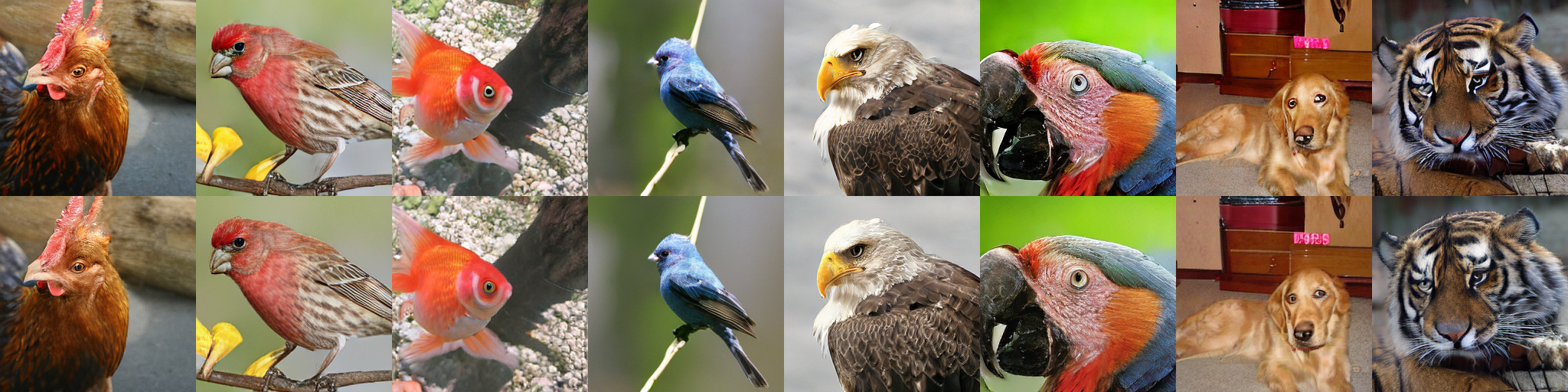}
\vspace{-3ex}
\caption{MeanFlow-XL/2: eight selected classes (top: before, bottom: after).}
\label{fig:qual-selected-meanflow}
\vspace{-4pt}
\end{figure}

\begin{figure}[!htbp]
\centering
\includegraphics[width=\linewidth]{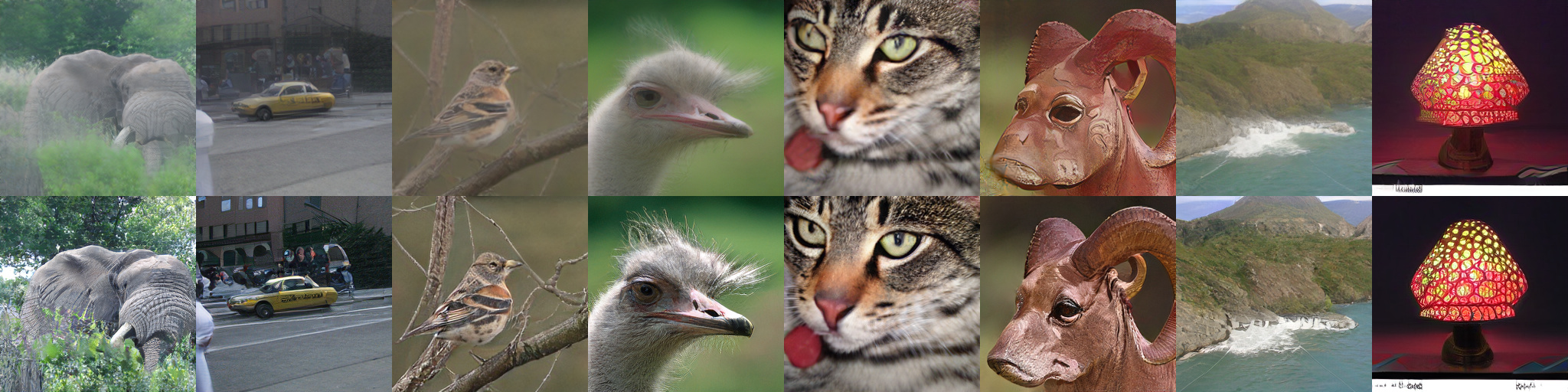}
\vspace{-3ex}
\caption{TiTok-L32: eight selected classes (top: before, bottom: after).}
\label{fig:qual-selected-titok}
\vspace{-4pt}
\end{figure}

Figures~\ref{fig:qual-selected-imf-m}--\ref{fig:qual-selected-titok} provide additional qualitative comparisons across iMF-M/2, MeanFlow-XL/2, DMF-XL/2+, and TiTok-L32. The top row (\emph{before}) shows samples generated using the pretrained decoder, while the bottom row (\emph{after}) shows samples generated using the LP-GANs-refined decoder, with corresponding samples sharing the same random seeds and class labels. These comparisons illustrate the visual improvements obtained through decoder refinement across different model scales and latent generation architectures.

\end{document}